\documentclass[10pt,twocolumn,letterpaper]{article}
\usepackage{graphicx}
\usepackage{amsmath}
\usepackage{amssymb}
\usepackage{booktabs}
\usepackage{enumitem}
\usepackage{verbatim}

\usepackage{subfigure}

\usepackage{wrapfig}
\usepackage{times}
\usepackage{epsfig}
\usepackage[utf8]{inputenc} % allow utf-8 input
\usepackage[T1]{fontenc}    % use 8-bit T1 fonts
\usepackage[pagenumbers]{cvpr} % To force page numbers, e.g. for an arXiv version
\DeclareMathOperator*{\argmax}{arg\,max}

\usepackage{siunitx}
\usepackage{cite}
\usepackage{ctable}
\usepackage{hhline}
\usepackage{booktabs}

\usepackage{csquotes}
\usepackage{pifont}
\usepackage{mathtools}
\usepackage{animate}

\usepackage{anyfontsize}
\usepackage[super]{nth}

\usepackage{ctable}
\usepackage{soul}
\usepackage{multicol}
\usepackage{makecell}
\usepackage[normalem]{ulem}
\usepackage{multirow}
\usepackage{float}
\usepackage{stfloats}

\usepackage{stmaryrd}
\usepackage{mathstyle}

\usepackage{color, colortbl}
\definecolor{Gray}{gray}{0.9}

\usepackage[pagebackref,breaklinks,colorlinks]{hyperref}

\newcommand{\modelname}{\textsc{LadderGNN}}

\usepackage[capitalize]{cleveref}
\crefname{section}{Sec.}{Secs.}
\Crefname{section}{Section}{Sections}
\Crefname{table}{Table}{Tables}
\crefname{table}{Tab.}{Tabs.}

\begin{document}

\title{Relational Graph Neural Network Design via Progressive Neural Architecture Search}

\newcommand*{\affaddr}[1]{#1} % No op here. Customize it for different styles.
\newcommand*{\affmark}[1][*]{\textsuperscript{#1}}
\newcommand*{\email}[1]{\texttt{#1}}
\makeatletter
\newcommand{\printfnsymbol}[1]{%
  \textsuperscript{\@fnsymbol{#1}}%
}
\makeatother
\author{%
Ailing Zeng\affmark[1], Minhao Liu\affmark[1], Zhiwei Liu\affmark[2], Ruiyuan Gao\affmark[1], Jing Qin\affmark[3], Qiang Xu\affmark[1]\\
\affaddr{\affmark[1]The Chinese University of Hong Kong},
\affaddr{\affmark[2]University of Illinois at Chicago},\\
\affaddr{\affmark[3]The Hong Kong Polytechnic University},
}

\maketitle

\vskip 0.3in

\begin{abstract}
We propose a novel solution to addressing a long-standing dilemma in the representation learning of graph neural networks (GNNs): how to effectively capture and represent useful information embedded in distant nodes to improve the performance of nodes with low homophily without leading to performance degradation in nodes with high homophily.
This dilemma limits the generalization capability of existing GNNs. 
Intuitively, interactions with distant nodes introduce more noise for a node than those with close neighbors. 
However, in most existing works, messages being passed among nodes are mingled together, which is inefficient from a communication perspective. 

Our solution is based on a novel, simple, yet effective aggregation scheme, resulting in a ladder-style GNN architecture, namely \modelname. 
Specifically, we separate messages from different hops, assign different dimensions for them, and then concatenate them to obtain node representations. 
Such disentangled representations facilitate improving the information-to-noise ratio of messages passed from different hops. 
To explore an effective hop-dimension relationship, we develop a conditionally progressive neural architecture search strategy. 
Based on the searching results, we further propose an efficient approximate hop-dimension relation function to facilitate rapid configuration of the proposed \modelname.
We verify the proposed \modelname~on seven diverse semi-supervised node classification datasets.
Experimental results show that our solution achieves better performance than most existing GNNs.
We further analyze our aggregation scheme with two commonly used GNN architectures, and the results corroborate that our scheme outperforms existing schemes in classifying low homophily nodes by a large margin. 
\end{abstract}


% \maketitle

\vspace{-0.2cm}
\section{Introduction}
Recently, a large number of research efforts have been dedicated to applying deep learning methods to graphs, known as graph neural networks (GNNs)~\cite{kipf2016semi,velivckovic2017graph}, achieving great success in modelling non-structured data, e.g., social networks~\cite{liu2020alleviating} and recommendation systems~\cite{liu2020basconv}. 

Learning an effective low-dimensional embedding to represent each node in the graph is arguably the most important task for GNN learning, wherein the node embedding is obtained by aggregating information with its direct and indirect neighbouring nodes passed through GNN layers~\cite{Gilmer2017NeuralMP}. 
Earlier GNN works usually aggregate with neighboring nodes that are within a short range (1-2 hops). 
For many graphs, this may cause the so-called \emph{under-reaching} issue~\cite{alon2020bottleneck} -- distant yet informative nodes are not involved, leading to unsatisfactory results, particularly when the homophily level of the graph is relatively low~\cite{pei2020geom}.

Consequently, lots of techniques that attempt to aggregate long-distant neighbours are proposed by deepening or widening the network~\cite{li2018deeper,xu2018powerful,xu2018representation, li2019deepgcns,zhu2021simple}. 
However, when we aggregate information from too many long-distant neighbours, the so-called \emph{over-smoothing} problem~\cite{li2018deeper} may occur, causing nodes to be less distinguishable from each other~\cite{chen2020measuring}.
To the end, the inference performance often greatly degrades, particularly at the nodes with high homophily level. 
This situation puts the GNN learning into a dilemma. 
On the one hand, to enhance the generalization capability of existing GNNs, particularly in dealing with graphs with low homophily, we should integrate information embedded in long-distance nodes in the learning process.
On the other hand, such an integration under existing aggregation schemes will often cause obvious performance degradation, particularly in graphs with high level of homophily~\cite{zhu2020beyond}.   

To alleviate this problem, several hop-aware GNN aggregation schemes are proposed in the literature~\cite{abu2019mixhop,zhanghop,zhu2019multi,wang2019heterogeneous,wang2021tree}. %\TODO{while those methods are still far from satisfactory. This work focuses on improving hop-aware aggregation for better node representation learning. Furthermore, we find that in existing hop-aware aggregation works, the messages passed among nodes are mingled together.} 
While showing promising results, the messages passed among nodes are mingled together in these approaches. From a communication perspective, mixing information from clean sources (mostly low-order neighbours) and that from noisy sources (mostly high-order neighbours) would inevitably cause difficulty for the receiver (i.e., the target node) to extract information.

\begin{figure*}[h]	
	\subfigure[Hop-1] %第一张子图
	{
		\begin{minipage}{3.7cm}
			\centering          %子图居中
			\includegraphics[width=\linewidth]{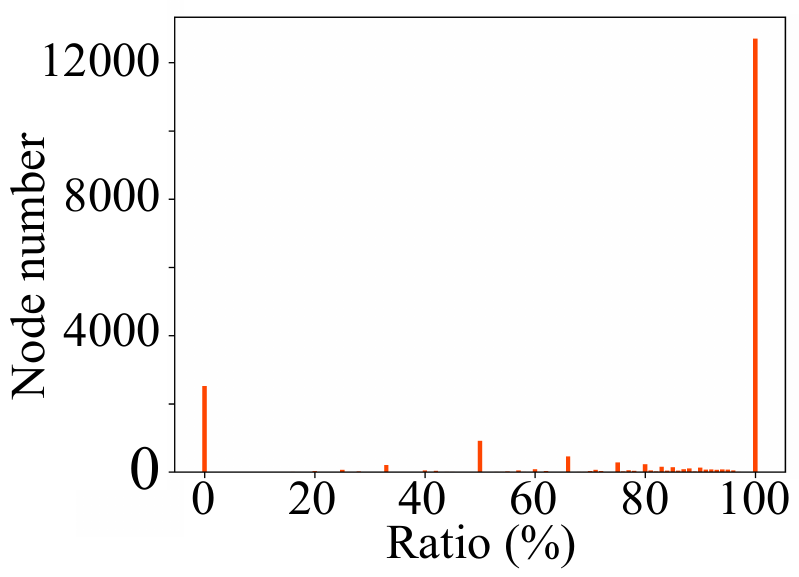}   %以pic.jpg的0.4倍大小输出
		\end{minipage}
	}
	\hspace{0.2cm}
	\subfigure[Hop-2] %第二张子图
	{
		\begin{minipage}{3.7cm}
			\centering      %子图居中
			\includegraphics[width=\linewidth]{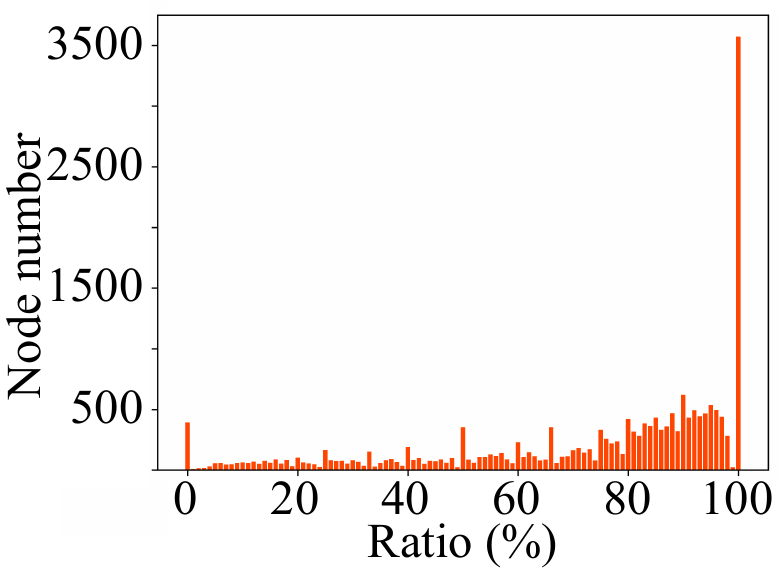}   %以pic.jpg的0.4倍大小输出
		\end{minipage}
	}
		\hspace{0.2cm}
	\subfigure[Hop-4] %第二张子图
	{
		\begin{minipage}{3.7cm}
			\centering      %子图居中
			\includegraphics[width=\linewidth]{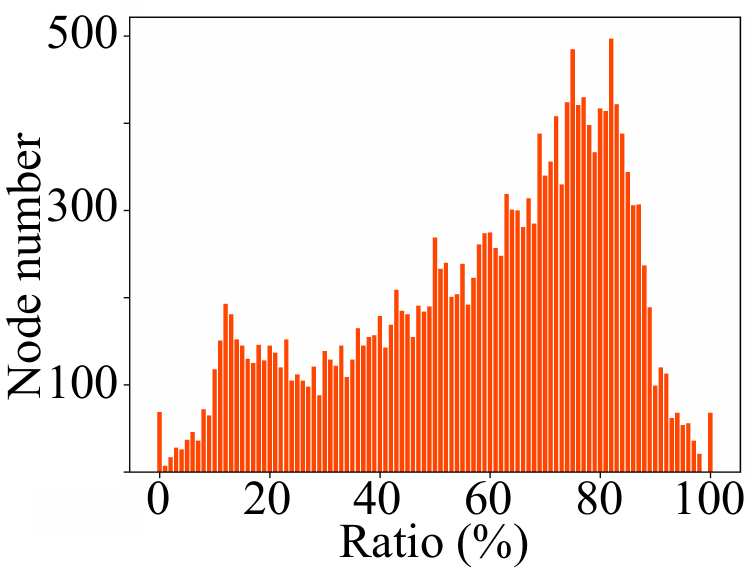}   %以pic.jpg的0.4倍大小输出
		\end{minipage}
	}
		\hspace{0.2cm}
	\subfigure[Hop-8] %第二张子图
	{
		\begin{minipage}{3.7cm}
			\centering      %子图居中
			\includegraphics[width=\linewidth]{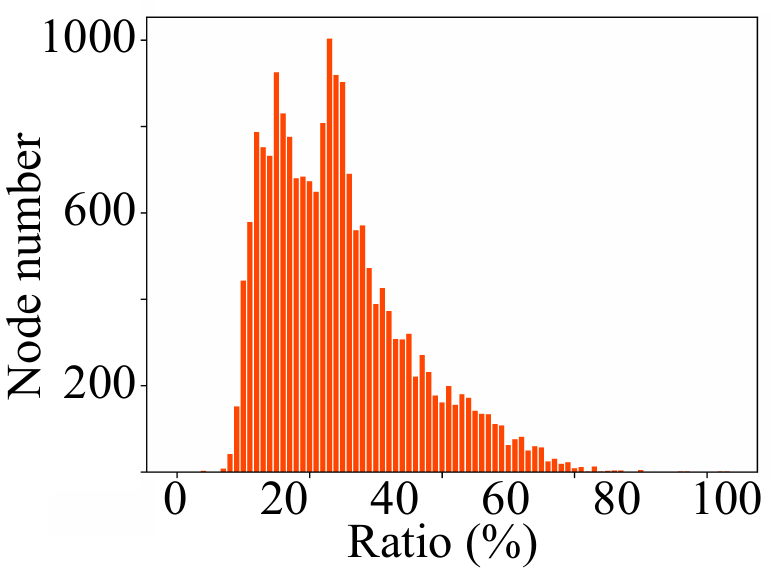}   %以pic.jpg的0.4倍大小输出
		\end{minipage}
	}
		\hspace{0.2cm}
	\vspace{-0.4cm}
	\caption{The histogram for the number of nodes ($y$ axis) with different homophily ratio, i.e., the percentage of neighbors with the same label as the target node at Hop-$k$ on the Pubmed dataset. It indicates a diminishing information-to-noise ratio for messages ranging from low-order hops to high-order hops. } %
	\label{fig:SNR}
	\vspace{-0.4cm}
\end{figure*}

Motivated by the above observations and considerations, we propose a novel, simple, yet effective hop-aware aggregation scheme, resulting in a ladder-style GNN architecture, namely \modelname, to comprehensively address this dilemma. 
The contributions of this paper include: 
% \vspace{-0.4cm}

\begin{itemize}
\item We take a communication perspective on GNN message passing. That is, we regard the target node for representation learning as the receiver and group the set of neighbouring nodes with the same distance to it as a transmitter that carries both information and noise. The dimension of the message can be regarded as the capacity of the communication channel. Then, aggregating neighbouring information from multiple hops becomes a multi-source communication problem with multiple transmitters over the communication channel.   
% \vspace{-0.5cm}
\item To improve node representation learning, we propose to separate the messages from different transmitters (i.e., Hop-$k$ neighbours), each occupying a proportion of the communication channel (i.e., disjoint message dimensions). As the homophily ratio in high-order neighbours is often lower than that in low-order neighbours, the resulting hop-aware representation is unbalanced with more dimensions allocated to low-order neighbours, leading to a ladder-style aggregation scheme.
% \vspace{-0.cm}

\item To explore the dimension allocation manner for neighbouring nodes with different hops effectively, we propose a conditionally progressive neural architecture search (NAS) strategy. Motivated by the search results,
we introduce an approximate hop-dimension relation function, which can generate close results to the NAS solution without applying compute-expensive NAS.
\vspace{-0.2cm}

\end{itemize}
% \vspace{-0.1cm}
To verify the effectiveness of the proposed GNN representation learning solution, we demonstrate it on seven semi-supervised node classification datasets for both homogeneous and heterogeneous graphs with different homophily levels. 
Experimental results show that our solution achieves better performance than most existing GNNs.
We further analyze our aggregation scheme with two commonly used GNN architectures, and the results corroborate that our scheme outperforms existing schemes in classifying low homophily nodes by a large margin.

\begin{figure*}[h]
\centering
\includegraphics[scale=0.3]{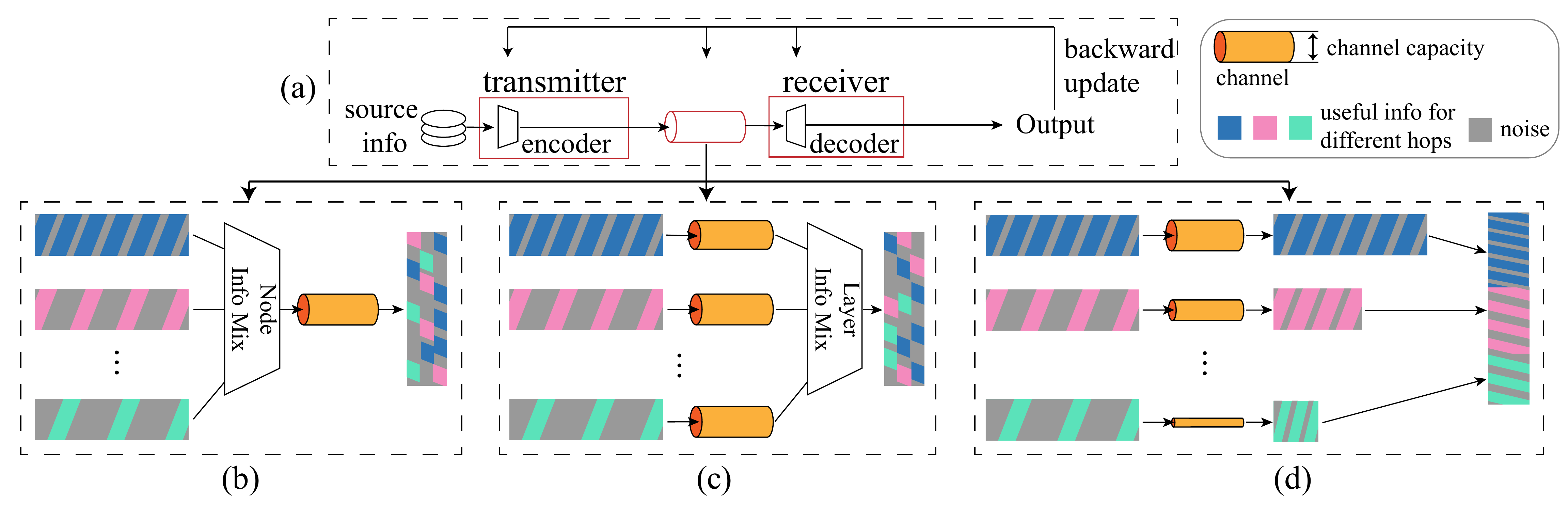}
\vspace{-0.2cm}
\caption{An illustration of GNN message passing and representation learning from a communication perspective. (a) A communication system contains transmitters that encode source information, communication channel, and receivers that decode the original information; (b) GNN representation learning with existing node aggregation scheme; (c) GNN representation learning with existing hop-aware aggregation scheme; (d) GNN representation learning with the proposed ladder-style aggregation scheme.}
\label{fig:channel}
\vspace{-0.4cm}
\end{figure*}

\vspace{-0.2cm}

\section{Related Work and Motivation}
\label{sec:relat}
\vspace{-0.1cm}
GNNs adopt message passing to learn node embeddings, which involves two steps for each node: neighbour aggregation and linear transformation~\cite{Gilmer2017NeuralMP}. The following formula presents the mathematical form of message passing in a graph convolutional network (GCN)~\cite{kipf2016semi}. Given an undirected graph $\mathcal{G}$= ($\mathcal{V}$, $\mathcal{E}$) with $N$ nodes and adjacency matrix $\mathbf{A}$, we can aggregate node features at the $l$-th layer $\mathbf{H}^{(l)} \in \mathbb{R}^{N \times C_i^{(l)}}$ as:
\vspace{-0.1cm}

\begin{equation}
    \mathbf{H}^{(l+1)} = \sigma(\Hat{\mathbf{A}} \mathbf{H}^{(l)} \mathbf{W}^{(l)}),
\end{equation}
where $\Hat{\mathbf{A}}=\mathbf{D}^{-1/2} (\mathbf{A}+\mathbf{I}_{N}) \mathbf{D}^{-1/2}$ is the augmented normalized adjacency matrix of $\mathcal{G}$. $\mathbf{I}_{N}$ is the identity matrix and $\mathbf{D}_{ii} = \sum_{j}(\mathbf{A}+\mathbf{I}_{N})_{ij}$. $\mathbf{W}^{(l)}\in \mathbb{R}^{C_i^{(l)} \times C_h^{(l)}}$ is the trainable weight matrix at the $l$-th layer used to update node embeddings. $C_i^{(l)}$ and $C_h^{(l)}$ are the channel size of the input and hidden layer, respectively. $\sigma(\cdot)$ is an activation function.

By taking the importance of different neighbours into consideration, graph attention network (GAT)~\cite{velivckovic2017graph} applies a multi-head self-attention mechanism during aggregation and achieves higher performance than GCN in many datasets. Recent GNN works improve GCN or GAT from two aspects. (i) Some make changes to the input graphs~\cite{rong2019dropedge,chen2020measuring,luo2021learning,zhao2020data,yang2021spagan}. For example,~\cite{luo2021learning} proposes to drop some task-irrelevant or noisy edges to achieve high generalization capability. 
The underlying thought of such changes is to increase the homophily level of the input graph so as to achieve satisfactory results even when aggregating a short range of neighboring nodes.
Unfortunately, for many practical applications, it is difficult, if not impossible, to design a high homophily graph. 
(ii) Without changing the graph, some GNN works try to further extract relevant information from high-order neighbours. Our work belongs to this category.
%and it can be easily combined with those GNN works in the first category that modifies the adjacency matrix $\mathbf{A}$. 

In order to aggregate information from high-order neighbours, some earlier works~\cite{li2019deepgcns,xu2018representation,xu2018powerful} simply stack deeper networks to retrieve such information recursively. To mitigate the possible over-fitting issues (caused by model complexity), SGC~\cite{wu2019simplifying} removes the nonlinear operations and directly aggregates node features from multiple hops. To relieve the potential over-smoothing problem that results in less discriminative node representations (due to over-mixing)~\cite{li2018deeper}, various hop-aware aggregation solutions are proposed. Some of them (e.g. HighOrder~\cite{morris2019weisfeiler}, MixHop~\cite{abu2019mixhop}, N-GCN~\cite{abu2020n}, GB-GNN~\cite{oono2020optimization}) employ multiple convolutional branches to aggregate neighbors from different hops. Others  (e.g. AM-GCN~\cite{wang2020gcn}, HWGCN~\cite{liu2019higher}, MultiHop~\cite{zhu2019multi}) try to learn adaptive attention scores when aggregating neighboring nodes from different hops. 
%\TODO{However, the performance of existing hop-aware methods still are not satisfactory.}

\iffalse
For example, MixHop concatenates the feature representations from multiple hops in a graph convolutional layer, which can be formulated as:
\begin{equation}
    \mathbf{H}^{(l+1)} = \parallel_{k \in (0, K]} \sigma(\Hat{\mathbf{A}}^k {\mathbf{H}}^{(l)} {\mathbf{W}}^{(l)}_k),
    \label{eq:mixhop}
\end{equation}
where $K$ indicates the number of hops used in aggregation. Accordingly, $\Hat{\mathbf{A}}^k$ means $k$ power of the normalized adjacency matrix $\Hat{\mathbf{A}}$. $\parallel$ denotes column-wise concatenation. $\mathbf{W}^{(l)}_k\in \mathbb{R}^{C_i^{(l)} \times C_h^{(l)}}$ is a learnable weight matrix of each hop $k$. By default, the output dimension $C_h^{(l)}$ of each branch is $1/K$ of the concatenated dimensions.
\fi

\iffalse
\begin{itemize}
    \item Both low-order neighbors and high-order neighbors contain useful information to the target node.
    \item High-order neighbors contain less useful information and more noise comparing to low-order neighbors.
\end{itemize}
\fi

% This work tries to address the dilemma in graph representation learning: aggregating long-range nodes to improve low homophily nodes and keep the performance on nodes with high homophily. We focus on the homophily at the node level to discover the degradation phenomena behind each target node. Therefore, 

As our scheme is closely related to the homophily level of a node, we provide a definition of homophily ratio and further explain our motivation based on this definition.
Given a graph $\mathcal{G} = (\mathcal{V},  \mathcal{E})$, we define the homophily ratio of the node $i$ as $r_i = \frac{\sum_{ j \in \mathcal{E}_{i,j}}(C(\mathcal{V}_{i})==C(\mathcal{V}_{j}))}{\sum_{ j \in \mathcal{E}_{i,j}}\mathcal{V}_{j}} $, where $j$ is its neighboring nodes, $C$ is the class label, and thus the $r_i$ indicates the percentage of nodes with the same class to $i$ in all its neighboring nodes. 
%The node with low homophily has a low ratio $r \rightarrow 0$, and vice versa.}

In Figure~\ref{fig:SNR}, we plot the homophily ratio for nodes in the Pubmed dataset with different hops. As can be observed from the figure, with the increase of hop distance, the percentage of neighbouring nodes with the same label decreases, indicating a diminishing information-to-noise ratio for messages ranging from low-order neighbours to high-order neighbours. 
This is a common phenomenon when the graph is well designed. 
Therefore, the critical issue in GNN message passing is how to retrieve information effectively while suppressing noise simultaneously. However, all the existing aggregation schemes do not explicitly and sufficiently consider this issue, making it challenging to achieve good performance in both high homophily nodes and low homophily nodes. This observation motivates us to propose our \modelname~architecture.

\vspace{-0.1cm}
\section{Method}
\label{sec:method}
\vspace{-0.1cm}

In Sec.~\ref{sec:motivation}, we take a communication perspective on GNN message passing and representation learning. Then, we give an overview of the proposed~\modelname~framework in Sec.~\ref{sec:method_la}. Next, we explore the dimensions of different hops with an RL-based NAS strategy in Sec.~\ref{sec:method_nas} and then introduce the approximate hop-dimension relation function in Sec.~\ref{sec:3.4}. 
%Finally, Sec.~\ref{sec:3.5} presents how to integrate our hop-aware aggregation scheme in GCN and GAT architectures.

\begin{figure*}[h]
    \centering
    % \vspace{-0.3cm}
    \includegraphics[scale=0.7]{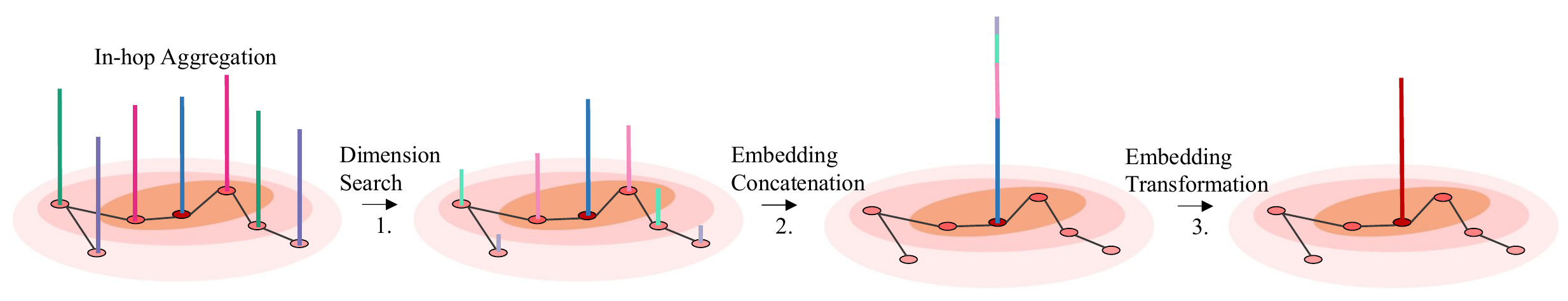}
    % \vspace{-0.5cm}
    \caption{Illustration of the process of \modelname~of one target node. The height of bars represents the dimension of embeddings from each node. Colors are with respect to hops.
    }
    \label{fig:frame}
    % \vspace{-0.5cm}
\end{figure*}

% \vspace{-0.3cm}

\subsection{GNN Representation Learning from a Communication Perspective}
\label{sec:motivation}
% \vspace{-0.1cm}

In GNN representation learning, messages are passed from neighbouring nodes to the target node and updated its embedding.
%, \TODO{which is similar to the principle of communication transmission.} Accordingly, we 
Figure~\ref{fig:channel} presents a communication perspective on GNN message passing, wherein we regard the target node as the receiver. Considering neighbouring nodes from different hops tend to contribute unequally
(see Figure~\ref{fig:SNR}), we group the set of neighbouring nodes with the same distance as one transmitter, and hence we have $K$ transmitters if we would like to aggregate up to $K$ hops. The dimension of the message can be regarded as the communication channel capacity. Then, GNN message passing becomes a multi-source communication problem.

Some existing GNN message-passing schemes  (e.g., SGC~\cite{wu2019simplifying}, JKNet~\cite{xu2018representation}, and S$^{2}$GC~\cite{zhu2021simple}) aggregate neighboring nodes before transmission, as shown in Figure~\ref{fig:channel}(b), which mix clean information source and noisy information source directly. The other hop-aware GNN message-passing schemes (e.g., AMGCN~\cite{wang2020gcn}, MultiHop~\cite{zhu2019multi}, and MixHop~\cite{abu2019mixhop}) as shown in Figure~\ref{fig:channel}(c)) first conduct aggregation within each hop (i.e., using separate weight matrix) before transmission over the communication channel, but they are again mixed afterward. 

Different from a conventional communication system that employs a well-developed encoder for the information source, one of the primary tasks in GNN representation learning is to learn an effective encoder that extracts useful information with the help of supervision. Consequently, mixing clean information sources (mostly low-order neighbours) and noisy information sources (mostly high-order neighbours) makes the extraction of discriminative features challenging. 

The above motivates us to perform GNN message passing without mixing up messages from different hops, as shown in Figure~\ref{fig:channel}(d). At the receiver, we concatenate the messages from various hops, and such disentangled representations facilitate extracting useful information from various hops with little impact on each other. Moreover, dimensionality significantly impacts any neural networks' generalization and representation capabilities~\cite{srivastava2014dropout,liu2016overfitting,alon2020bottleneck,bartlett2020benign}, as it controls the amount of quality information learned from data. In GNN message passing, the information-to-noise ratio of low-order neighbours is usually higher than that of high-order neighbours. Therefore, we tend to allocate more dimensions to close neighbours than distant ones, leading to a ladder-style aggregation scheme.

% \vspace{-0.3cm}
\subsection{Ladder-Aggregation Framework}
\label{sec:method_la}
% \vspace{-0.1cm}
% 为了更好地解决高阶邻居聚合的问题，我们提出维度控制的方法。
With the above, Figure~\ref{fig:frame} shows the node representation update procedure in the proposed \modelname~architecture. For a particular target node (the center node in the figure), we first aggregate node within each hop, which can be conducted by existing node-wise aggregation methods (e.g., GCN or GAT). Next, we determine the dimensions for the aggregated messages from different hops and then concatenate them, instead of mixing them up, for inter-hop aggregation. Finally, we perform a linear transformation to generate the updated node representation.

Specifically, $K$ is the maximum number of neighboring hops for node aggregation. For each group of neighbouring nodes at Hop-$k$, we determine their respective optimal dimensions and then concatenate their embeddings into $\mathbf{H}$ as follows:
% \vspace{-0.2cm}

\begin{equation}
    \mathbf{H} =\parallel_{k \in [1, K]} (\widehat{\mathbf{A}}_k \mathbf{X} \mathbf{W}_k),
    \label{eq:low_dim3}
\end{equation} 
% \vspace{-0.2cm}

where $\widehat{\mathbf{A}}_k$ is the normalized adjacency matrix of the $k_{th}$ hop and $\mathbf{X}$ is the input feature. A learnable matrix $\mathbf{W}_k\in \mathbb{R}^{C_{i} \times C_{o,k}}$ controls the output dimension of the $k_{th}$ hop as $C_{o,k}$. $\parallel$ means concatenation.
Encoding messages from different hops with distinct $\mathbf{W}_k$ avoids the over-mixing of neighbours, thereby alleviating the impact of noisy information sources on clean information sources during GNN message passing. Accordingly, $\mathbf{H}$ is a hop-aware disentangled representation of the target node. Then, with the classifier $f$ after the linear layer $\mathbf{W}_U$, we have:
% \vspace{-0.2cm}

\begin{equation}
    \mathbf{\widehat{Y}} = f(\mathbf{H} \mathbf{W}_U),
    \label{eq:low_dim1}
\end{equation}
% \vspace{-0.2cm}

where $\mathbf{\widehat{Y}}$ is the output softmax values. Given the supervision $\mathbf{Y}$ of some nodes, we can use a cross-entropy loss to calculate gradients and optimize the above weights in an end-to-end manner.

With the above, if the adjacency matrix $\mathbf{A}$ are the same as the original GCN architecture, the resulting GNN architecture with our ladder-aggregation framework is namely \emph{Ladder-GCN}. Similarly, when we employ a self-attention scheme within hops to obtain the attention-based adjacency matrix $\Hat{\mathbf{A}}$ as in the original GAT architecture, the resulting GNN architecture is namely \emph{Ladder-GAT}. Please note, our proposed ladder-aggregation scheme could also be integrated into other GNN architectures (e.g.,~\cite{luo2021learning}).

% \vspace{-0.3cm}
\subsection{Hop-Aware Dimension Search}
\label{sec:method_nas}
% \vspace{-0.1cm}

% If dimensionality is not set properly in representation learning, under-fitting or over-fitting may occur~\cite{murphy2012machine}. Hence,
Allocating different dimensions for messages from different hops is the key in \modelname~design. As there are numerous hop-dimension allocation possibilities, determining an appropriate allocation is a non-trivial task. 

In recent years, neural architecture search~(NAS) has been extensively researched, which automatically designs deep neural networks with comparable or even higher performance than manual designs by experts (e.g.,~\cite{Bello2017NeuralOS, tan2019mnasnet, zoph2018learning, liu2018darts,Liu_2019_ICCV}).
Existing NAS works in GNNs~\cite{ijcai20gao,Zhou2019AutoGNNNA,shi2020evolutionary} search the graph architectures (e.g., \emph{1-hop} aggregators, activation function, aggregation type, attention type, etc) and hyper-parameters to reach better performance. However, they ignore to aggregate multi-hop neighbours, let alone the dimensionality of each hop. In the following, we introduce our proposed NAS solution.

\textbf{Search Space:}~Different from previous works in GNNs~\cite{ijcai20gao,Zhou2019AutoGNNNA,you2020design}, our search space focuses on the dimension of each hop, called hop-dimension combinations.
To limit the possible search space $\mathcal{O}$ for hop-dimension combinations, %for hop $k$,determine an optimized hop-dimension 
we apply exponential sampling {$2^0$, $2^1$, $2^2$,...,$C_i$,...,$2^{(n-1)}$, $2^{(n)}$} strategies for dimensions. $n$ are hyper-parameters, representing the index and sampling granularity to cover the possible dimensions. For each strategy, the search space should also cover the dimension of initial input feature $C_i$.

\begin{figure}[t]
\centering
\includegraphics[scale=0.45]{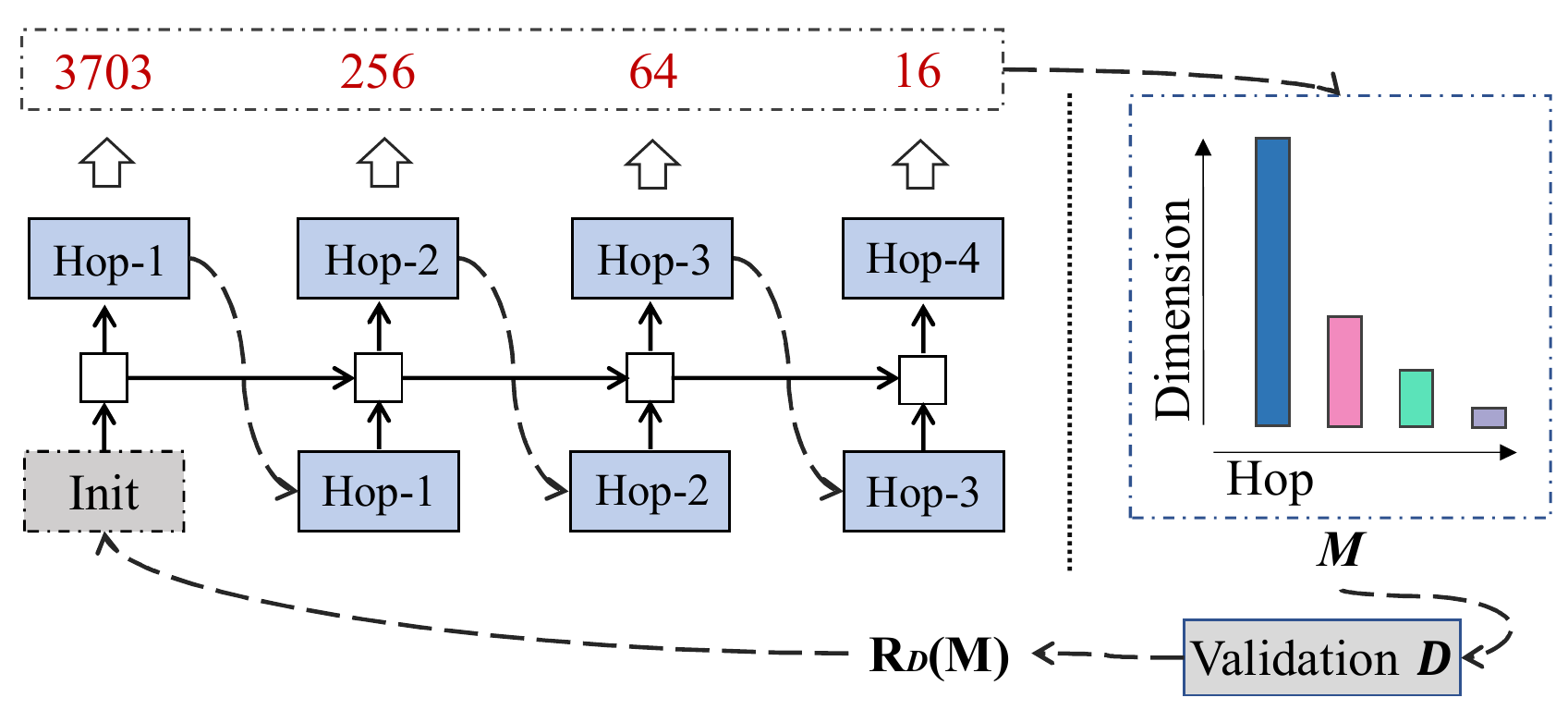}
% \vspace{-0.3cm}
\caption{An illustration of RL-based NAS for the hop-dimension exploration. A recurrent network (controller) generates descriptions of the dimension for each hop. Once the controller generates a framework $M$, it will be trained on the training set and then tested on the validate set $D$. The validation result $R_D(M)$ is taken as the reward to update the recurrent controller.}
% \vspace{-0.5cm}
\label{fig:nas}
\end{figure}

\textbf{Basic Search Algorithm:}
% \label{sec:method_sa}
Given the search space $\mathcal{O}$, we target finding the best model $M^{*} \in \mathcal{M}$ to maximize the expected validation accuracy. We choose
the reinforcement learning strategy since its reward is easy to customize for our problem. As shown in Figure \ref{fig:nas}, a LSTM controller based on the parameters $\theta$ generates a sequence of actions $a_{1:K}$ with length $K$, where each hop dimension $C_k$ ($1 \leq k \leq K$) is sampled from the search space mentioned above. Then, we can build a model $M$ mentioned in Sec.~\ref{sec:method_la}, and train it with a cross-entropy loss function. Then, we test it on the validation set $D$ to get an accuracy $R_D(M)$. Next, we can use the accuracy as a reward signal and perform a policy gradient algorithm to update the parameters $\theta$ so that the controller can generate better hop-dimension combinations iteratively.
The objective function of the model is shown in:
\begin{equation}
    M^{*} = \argmax_{M} \; \mathbb{E}_{P(a_{1:K}, \theta)}[R_D(M)].
    \label{eq:obj}
\end{equation}

\textbf{Conditionally Progressive Search Algorithm:}
Considering the extremely large search space with the basic search algorithm, e.g., the search space size will be $(n+1)^{k}$ for the exponential sampling with $k$ hops. This makes it more challenging to search for the optimal combinations with limited computational resources. Moreover, we find that there are a large number of redundant actions in our search space. To improve the efficiency and effectiveness of the search procedure, we are inspired to propose a conditionally progressive search algorithm.

That is, instead of searching the entire space all at once, we divide the searching process into multiple phases, starting with a relatively small number of hops, e.g., $K = 3$. After obtaining their results, we only keep those hop-dimension combinations that are promising, where they are regarded as the conditional search space, with high $R_D(M)$.

Next, we conduct the hop-dimension search for the ($K$+$1$)$_{th}$ hop based on the conditional search space filtered from the last step, and again, keep those combinations with high $R_D(M)$. This procedure is conducted progressively until aggregating more hops cannot boost performance. With this algorithm, we can largely reduce the redundant search space to enhance search efficiency.

% \vspace{-0.1cm}
% \subsection{The Approximate Hop-Dimension Relation Function}
\subsection{Hop-Dimension Relation Function}
\label{sec:3.4}
% The computational resources required to conduct NAS are extremely expensive, even with the proposed progressive search algorithm. Moreover, after analyzing the hop-dimension combinations in Sec.~\ref{sec:homo_nas}, we find that most of the satisfactory combinations show rather consistent principle.
% \vspace{-0.1cm}
The computational resources required to conduct NAS are extremely expensive for large graphs, even with the proposed progressive search algorithm. Therefore, it is essential to have an efficient method to determine the dimensionality of every hop in practical applications.
From our NAS experimental results (Sec.~\ref{sec:homo_nas}), we observe that the low-order neighbours within $L$ hops are usually directly aggregated with the original feature dimensions while high-order neighbours are associated with an approximately \emph{exponentially decreasing} dimensions. 

This motivates us to propose a simple yet effective \emph{hop-dim relation function} to approximate the NAS solutions. The output dimension of $k_{th}$ hop is:

\begin{equation}
    C_{o,k} = d^{\max{\{k-L,0\}}}*C_i,
    \label{eq:dim1}
\end{equation}
where $0<d<1$ is the dimension compression ratio, and $C_i$ is the dimension of the input feature. With such an approximate function, there is only one hyper-parameter to determine, significantly reducing the computational cost. 
\section{Experiment}
\label{sec:exp}
% \vspace{-0.1cm}
In this section, we validate the effectiveness of \modelname~on seven widely-used semi-supervised node classification datasets. We first analyze the NAS results in Sec.~\ref{sec:homo_nas}. Then, we combine the proposed hop-aware aggregation scheme with the approximate function with existing GNNs in Sec.~\ref{sec:high}. Moreover, as a new hop-aware aggregation scheme, in Sec.~\ref{sec:hop_results}, we quantitatively compare with exiting works. Furthermore, we conduct experiments on heterogeneous graphs in Sec.~\ref{sec:hete_exp}. Last, we show an ablation study on the proposed hop-dim relation function in Sec.~\ref{sec:homo_ab}.

\textbf{Data description:} For the semi-supervised node classification task on homogeneous graphs, we evaluate our method on \textbf{five} datasets: Cora~\cite{yang2016revisiting}, Citeseer~\cite{yang2016revisiting}, Pubmed~\cite{yang2016revisiting}, OGB-Arxiv~\cite{hu2020open} and OGB-Products~\cite{hu2020open}. We split the training, validation and test  set following earlier works~\cite{abu2019mixhop,wang2020gcn,kipf2016semi,velivckovic2017graph}. Furthermore, on heterogeneous graphs, we verify the methods on two datasets: ACM and IMDB~\cite{wang2019heterogeneous}. Due to page limits, more details about \emph{dataset descriptions}, \emph{data pre-processing} procedure and \emph{more comparison with existing methods} are listed in the Appendix. 

%wherein we differentiate attributes and aggregates them in a similar disentangled manner. 

\begin{figure}[t!]
% \begin{figure}
    \centering
    \includegraphics[scale=0.55]{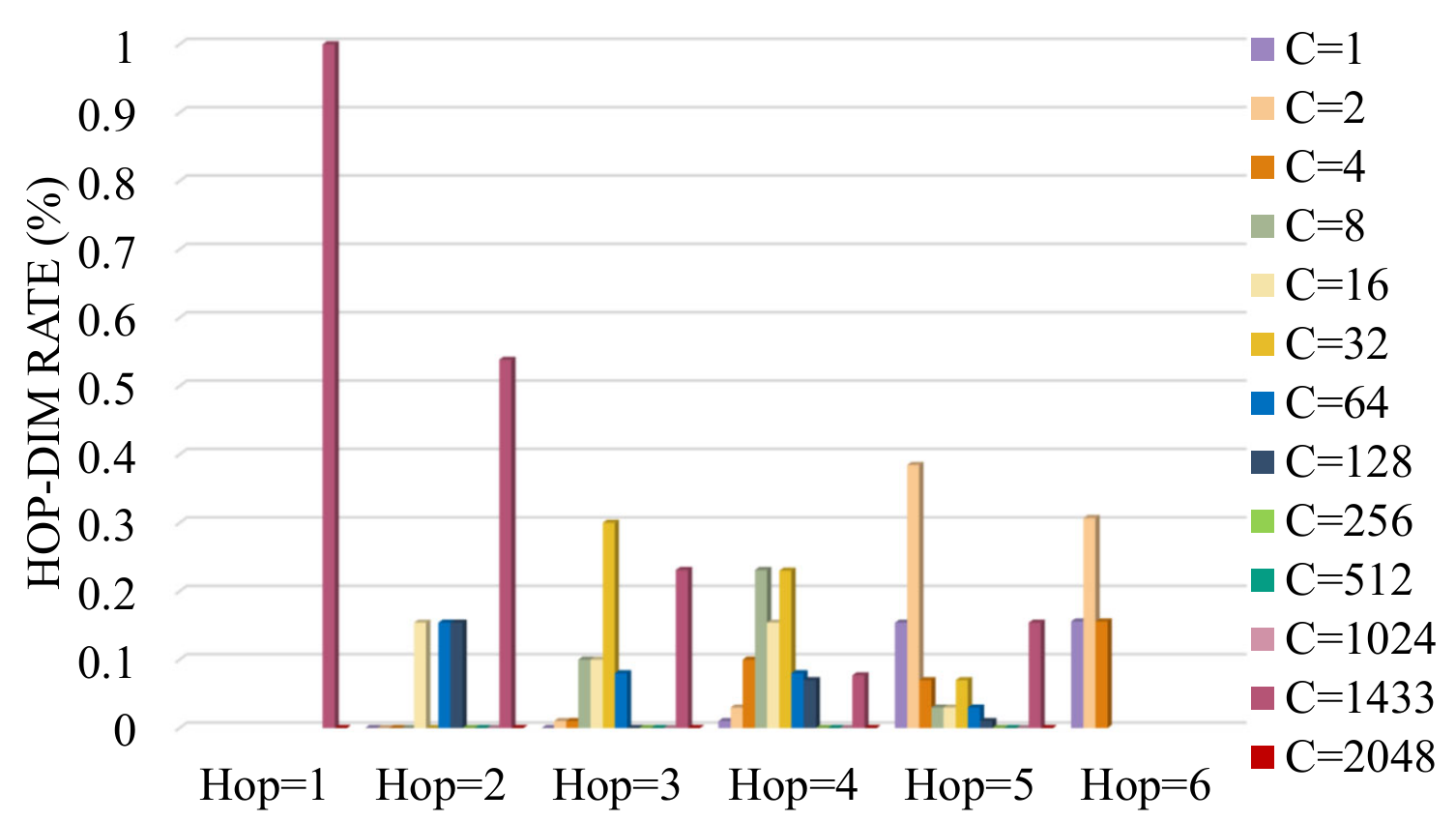}
    % \vspace{-0.3cm}
    \caption{The histogram of the filtered dimensions for different hops on the Cora dataset. X-axis illustrates Hop-$k$ with potential dimension assignment, while Y-axis shows the rate (\%) of the dimension for the corresponding hop. The colored bars represent the size of different dimensions, as shown in the upper legend. }
    \label{fig:nas_vis}
    % \vspace{-0.3cm}
\end{figure}

% \vspace{-0.1cm}
\subsection{Results from Neural Architecture Search}
\label{sec:homo_nas}
% \vspace{-0.1cm}
%The former one utilizes RNN to sample the whole neural architecture, and applies reinforcement rule to update controller.% with the previous NAS approaches based on kinds of strategies
To study the impact of the dimensions among hops, we conduct the NAS on different datasets to find out the optimal hop dimension combinations.
There exist a number of NAS approaches for GNN models, including random search (e.g.,~AGNN~\cite{Zhou2019AutoGNNNA}), reinforcement learning-based (RL) solution (e.g.,~GraphNAS~\cite{ijcai20gao} and AGNN~\cite{Zhou2019AutoGNNNA}) and evolutionary algorithm (e.g.,~Genetic-GNN~\cite{shi2020evolutionary}), wherein the RL-based solutions are more effective than others. Thus, in this work, we follow the same strategy as RL-based solutions to search for appropriate dimensions allocated to each hop.
% Hence, we can design an approximate method for simplification.

\begin{table}[htbp]
% \vspace{-5mm}
\scriptsize
\caption{The accuracy(\%) comparison for different NAS strategies. Ours-with \emph{cond.} in our method means using the conditional search strategy. Ours- \emph{Approx.} indicates the results is obtained from the hop-dimension relation function. The best result is in bold. The second-place result is underlined. \emph{share} means to train the architecture with shared weight.}
\label{tab:nas}
\begin{center}

{
\begin{tabular}{ l |c |c|c}
\hline
Method&Cora&Citeseer&Pubmed\\
\hline
Random-w/o share~\cite{Zhou2019AutoGNNNA} &81.4&72.9&77.9\\
Random-with share~\cite{Zhou2019AutoGNNNA}&82.3&69.9&77.9\\
GraphNAS-w/o share~\cite{ijcai20gao} &82.7&73.5&78.8\\
GraphNAS-with share~\cite{ijcai20gao} &83.3&72.4&78.1\\
AGNN-w/o share~\cite{Zhou2019AutoGNNNA} &\underline{83.6}&73.8&79.7\\
AGNN-with share~\cite{Zhou2019AutoGNNNA} &82.7&72.7&79.0\\
Genetic-GNN~\cite{shi2020evolutionary} &\textbf{83.8}&73.5&79.2\\
\hline
Ours-w/o \emph{cond.}&82.0&72.9&79.6\\
Ours-w/ \emph{cond.}&83.5&\textbf{74.8}&\textbf{80.9}\\
Ours-\emph{Approx.} &83.3&\underline{74.7}&\underline{80.0}\\
\hline
% \vspace{-1cm}
\end{tabular}}
\end{center}
\end{table}

In particular, we search the hop-dimension combinations of $10$ hops on Cora, Citeseer, and Pubmed datasets and show experimental results in Table~\ref{tab:nas}. Compared with existing NAS methods, our NAS method achieves better results with conditional progressive search algorithm on Citeseer and Pubmed datasets, improving over \emph{Genetic-GNN} by 1.4\% and 2.1\%, respectively. Meanwhile, we achieve comparable accuracy in the Cora dataset only by considering the hop-dimension combinations. Moreover, compared with \emph{w/o cond.}, we can find 2.6\% improvements on conditional progressive search, indicating the effectiveness of this strategy to search optimal hop dimension combinations under a limited search resource. Moreover, the \emph{Approx.} method show competitive results with NAS-based results, especially on Cora and Citeseer datasets.

Specially, we demonstrate the histogram of the possible dimension assignment for different hops in Figure~\ref{fig:nas_vis}. We can obtain two observations: (i) for low-order neighbors, i.e., when hop is less than $3$ in this case, most of the sorted solutions with high accuracy keep the initial feature dimension;
(ii) most of the possible dimensions of the hop are only in single digits, which verify the necessity of the proposed conditional strategy to reduce the search space greatly; 
(iii) The dimensionality tends to be reduced for high-order neighbours, and approximating it with exponentially decreasing dimensions occupies a relatively large proportion of the solutions.

% \red{Although the NAS can achieve superior results than other baselines, it requires high computational resources. Therefore, the proposed approximate hop-dimension relation function inspired by NAS results can achieve comparable performance but is more efficient with only one hyper-parameter.}

Last, the above results serve two purposes: (i) they facilitate and support the design of the proposed approximate hop-dimension relation function; (ii) they validate the effectiveness of the proposed approximate hop-dimension relation function. Accordingly, we could use the approximate relation function with only one parameter to search for proper hop dimensions with comparable performance to the NAS solution.

% \vspace{-0.3cm}
\subsection{Comparison with General GNNs}
\label{sec:general}
% \vspace{-0.1cm}
\textbf{Quantitative Results:} We demonstrate the accuracy among general GNNs, like GCN, GAT and GraphSage on five popular datasets in Table~\ref{tab:compare}. As an effective hop-aware aggregation, we take GCN and GAT as examples to integrate them into our aggregation framework. The results show our methods can boost the performance of GCN and GAT by up to 4.7\%, indicating the proposed aggregation is beneficial and robust on different dataset.

\begin{table}[h]
\caption{GNN performance comparison.}
\scriptsize
\begin{center}
{
\begin{tabular}{ ll |c|c |c|c|c}
\hline

Method&Cora&Citeseer&Pubmed&Arxiv&Products&Average\\
\hline
% \multirow{6}{*}{\rotatebox[origin=c]{90}{Average Acc. (\%)}}

GCN~&81.5&70.3&79.0& 71.7  & 75.6&75.6\\
GraphSage&81.3&70.6&75.2& 71.5  & 78.3&75.4\\
GAT&78.9&71.2&79.0 & \underline{73.6}  & \underline{79.5}&76.4\\
SGC&81.0&71.9&78.9& 68.9  & 68.9&73.9\\
\hline
Ladder-GCN&\textbf{83.3}&\textbf{74.7}&\underline{80.0}& 72.1  & 78.7 &\underline{77.8}\\
Ladder-GAT&\underline{82.6}&\underline{73.8}&\textbf{80.6}& \textbf{73.9} & \textbf{80.8} &\textbf{78.3}\\

\hline
\end{tabular}}
\end{center}
\label{tab:compare}
\end{table}

\begin{figure}[h]	
\centering
    \subfigure[\emph{Citeseer} dataset.] 
	{
		\begin{minipage}[t]{0.32\linewidth}
			\centering      
			\includegraphics[width=1.6in]{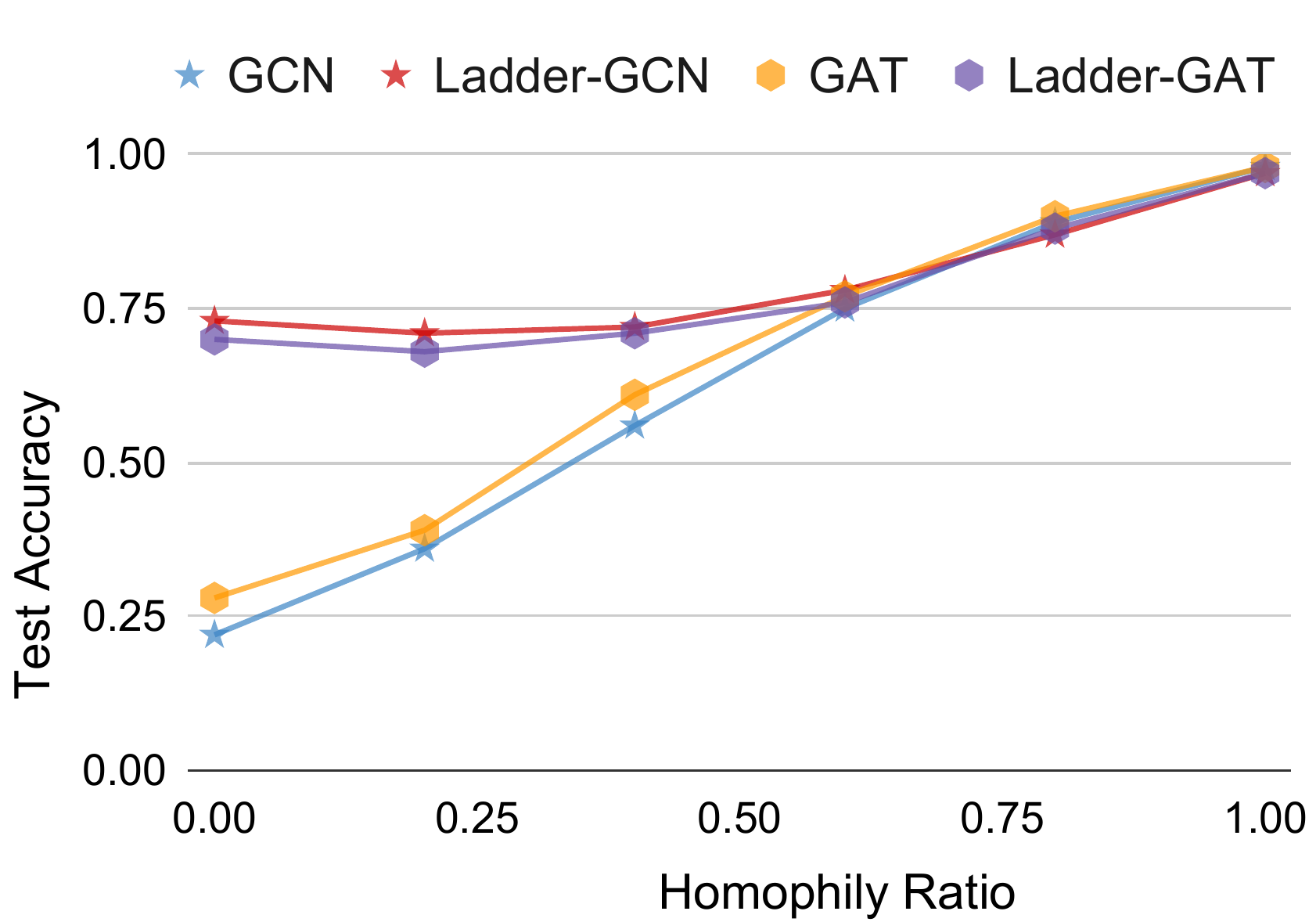}
		\end{minipage}
	}
    	\label{fig:cite_homo}  
    	\hspace{1cm}
	\subfigure[\emph{Cora} dataset.] 
	{
		\begin{minipage}[t]{0.32\linewidth}
			\centering         
			\includegraphics[width=1.6in]{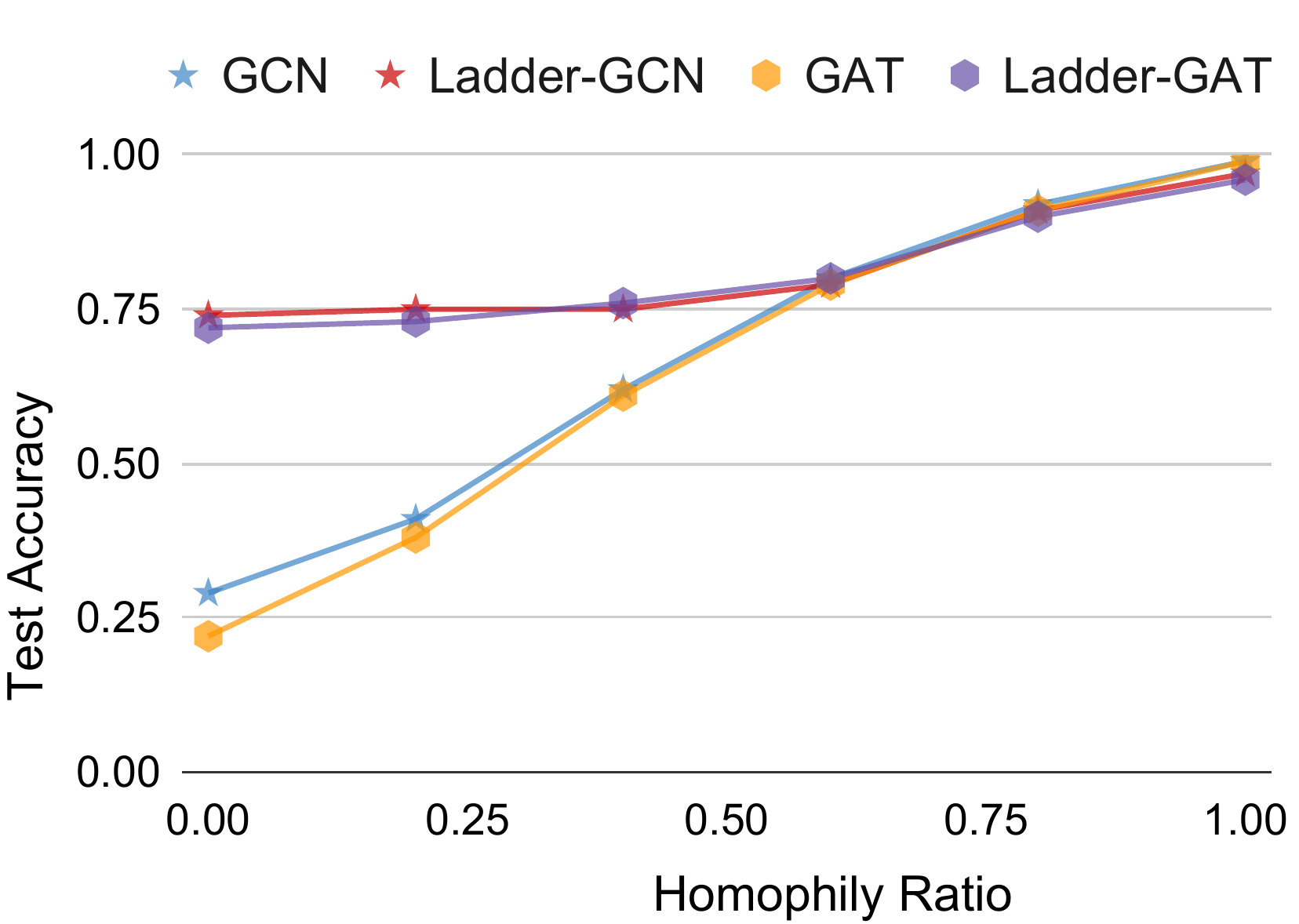}   
		\end{minipage}
	} 
    	\label{fig:cora_homo} 
    	\hspace{1cm}
% \vspace{-0.2cm}
\caption{Results under different homophily ratios for the Citeseer and Cora datasets. \modelname~outperforms GCN and GAT by a large margin when classifying low homophily nodes, while keeping competitive performance on high homophily nodes.}
% \vspace{-0.4cm}
\label{fig:homo} 
\end{figure}

\textbf{Detailed Analysis:}
To understand the benefits provided by \modelname, in Figure 6, we show the classification accuracy for nodes under different homophily ratio. As expected, high homophily nodes (e.g., when the ratio is higher than 50\%) are relatively easy to classify with high accuracy, and there is not much difference between the results of our approach and those of existing methods. For the low homophily nodes (e.g., when the ratio is smaller than 25\%), \modelname~clearly excels with much higher classification accuracy and it even sustains at similar accuracy as those nodes with homophily ratio of 60\%.    

%To explore the effectiveness of our method, we compare the accuracy of the nodes at different homophily ratios under the most used baselines (GCN and GAT) with the proposed aggregation scheme. In Figure~\ref{fig:homo}, we find that our method surpasses the basic models by about $50$\% at $0$ to $0.2$ homophily ratios consistently. With the increase of homophily ratios, the accuracy of these methods tends to be saturated. Thus, these results can verify the generalization ability of \modelname under different homophily ratios. As most graphs have larger homophily ratios, our method does not appear to improve as much in the final average accuracy. Our method can be prominent on data with low homophily ratios, and relieve the dilemma.

\begin{table}[h]
% \vspace{-0.5cm}
\caption{The comparison of efficiency on Citeseer dataset.}
% \scriptsize
\label{tab:efficiency}
\begin{center}
% \scalebox{0.5}
% \resizebox{\textwidth}{mm}
\tiny
\begin{tabular}{ l |c |c|c|c|c}
\hline
Methods	&GCN	&GAT	&SGC&	MixHop&Ours\\
\hline
Accuracy (\%)&70.3$\pm$0.2&71.2$\pm$0.7&71.9$\pm$0.5&73.0$\pm$0.1&	\textbf{74.7$\pm$0.3}\\
Parameter(K)	&118.7 &237.5&\textbf{22.2}&177.9&	107.6\\
Train Time(ms)&5.9$\pm$1.0&525.0$\pm$5.5&\textbf{1.2$\pm$0.4}&4.9$\pm$0.3&2.0$\pm$0.5\\
Test Time(ms)&	9.8$\pm$0.8&181.4$\pm$0.5&\textbf{4.6$\pm$0.3}	&10.5$\pm$2.43&6.2$\pm$0.9\\
Memory(MiB)&\textbf{723.0}&10,565.0&753.0&789.0&779.0\\
\hline
\end{tabular}
\end{center}
\end{table}

\textbf{Efficiency Exploration:}
% \label{sec:effic}
% Computational efficiency is an essential factor in GNN design. Hence, it is preferred not to use NAS due to its high computational cost, especially in real-life scenes. Hence, when performing the comparative study with existing solutions beyond Section~\ref{sec:homo_nas}, we use the proposed approximate hop-dimension relation function (instead of NAS) to determine the dimension allocation for neighboring nodes from different hops in \modelname. In other words, the advantages of \modelname~ are achieved without applying compute-expensive NAS.
A comparison of our \emph{Ladder-GCN} model with other related works (GCN, SGC, GAT, and MixHop) is shown in Table~\ref{tab:efficiency}. Experimental results are performed on a TITAN-Xp machine. We average ten runs on the Citeseer dataset to obtain the average training time for each epoch, the average test time, and the model accuracy. The settings of the other methods follow the respective papers, and we set $K$ = $5$ and $d$ = $0.0625$ in \emph{Ladder-GCN}. As can be seen from Table~\ref{tab:efficiency}, the computational time and memory costs of \emph{Ladder-GCN} are moderately larger than SGC, with higher model accuracy.

\begin{table}[h]
\scriptsize
\caption{Comparison with Hop-aware GNNs.}
\begin{center}
{
\begin{tabular}{ l |l|c |c|c}
\hline
% \multicolumn{2}{c|}{Method}&Cora&Citeseer&Pubmed\\
Method&Cora&Citeseer&Pubmed&Average\\
\hline
% \multirow{11}{*}{\rotatebox[origin=c]{90}{Average Accuracy (\%)}}
HighOrder~\cite{morris2019weisfeiler}&76.6&64.2&75.0&71.9\\
MixHop~\cite{abu2019mixhop}&80.5&69.8&79.3&76.5\\
GB-GNN~\cite{oono2020optimization}&80.8&70.8&79.2&76.9\\
HWGCN~\cite{liu2019higher}&81.7&70.7&79.3&77.2\\
MultiHop~\cite{zhu2019multi}&82.4&71.5&79.4&77.8\\
HLHG~\cite{lei2019hybrid}&82.7&71.5&79.3&77.8\\
AM-GCN~\cite{wang2020gcn}&82.6&\underline{73.1}&79.6&\underline{78.4}\\
N-GCN~\cite{abu2020n}&83.0&72.2&79.5&78.2\\
TDGNN-w~\cite{wang2021tree} &\underline{83.4}&70.3&\underline{79.8}&77.8\\
TDGNN-s~\cite{wang2021tree} &\textbf{83.8}&71.3&\textbf{80.0}&\underline{78.4}\\
\hline
Ladder-GCN&83.3&\textbf{74.7}&\textbf{80.0}&\textbf{79.3} \\
\hline
\end{tabular}}
\end{center}
\label{tab:hop_results}
% \vspace{-0.3cm}
\end{table}

% \vspace{-0.1cm}
\subsection{Comparison with Hop-aware GNNs}
\label{sec:hop_results}
% \vspace{-0.1cm}
%This work focuses on exploring  to explore how powerful hop-aware aggregation should be. 
Our proposed solution improves the hop-aware GNN representation learning capability when aggregating high-order neighbors during message passing. Table~\ref{tab:hop_results} presents the comparison with other hop-aware solutions. 
%\newpage

As can be observed, in terms of Top-1 accuracy (\%), 
%the best existing hop-aware models for the three datasets are different, while 
our \modelname~with a simple hop-dimension relation function achieves competitive results on all three datasets, indicating the effectiveness of the proposed solution. Specifically, our method shows more improvements on Citeseer than on Cora and Pubmed. We attribute it to the fact that CiteCeer has a relatively lower graph homophily ratio of $0.74$ while the homophily ratios of Cora and Pubmed are $0.81$ and $0.80$, respectively. Because our \modelname~can extract discriminative information from noisy graph, the improvements on Citeceer is more significant. 

%According to ~\cite{zhu2020beyond}, Unlike the high graph homophily rate of Cora ($0.81$) and Pubmed ($0.80$), Citeseer has a lower graph homophily rate as $0.74$, making high-order neighbours more noisy and hard to extract discriminative information. Consequently, the fact that our method can boost relatively more on Citeseer proves our method's effectiveness in handling noisy graphs. }

%Noted the results from \emph{Ours} are conducted with the proposed \emph{hop-dim relation function}.

% \vspace{-0.2cm}
\subsection{Heterogeneous Graph Representation Learning}
\label{sec:hete_exp}
% \vspace{-0.1cm}
Heterogeneous graphs that consist of different types of entities (i.e., nodes) and relations are ubiquitous.
In heterogeneous graph representation learning,  meta-path-based solutions are proposed to model the semantics of different relations among entities, e.g., Movie-Actor-Movie and Movie-Director-Movie in the IMDB dataset.

\begin{table*}[h]
%\vspace{-5mm}
\scriptsize
\begin{center}
\caption{Comparison of the accuracy (\%) of the two methods on two datasets (Higher values are better). GAT and HAN are re-implemented by us to keep all hyper-parameters as the same. \emph{All} uses all meta-paths. \emph{IMP.} is an abbreviation of improvement.}
\label{tab:hete_att}
%\scalebox{0.75}
{
\begin{tabular}{ l |c |c|c|c|c|c|c|c}
\hline
Dataset&\multicolumn{4}{c|}{ACM}&\multicolumn{4}{c}{IMDB}\\
\hline
Method&PAP$\&$PSP&PAP$\&$PTP&PSP$\&$PTP&\emph{All} &MAM$\&$MDM&MAM$\&$MYM&MDM$\&$MYM&\emph{All}\\
\hline
GAT&83.5&79.7&79.7&79.7&53.0&40.9&48.9&48.2\\
HAN&87.9&84.9&82.1&88.7&52.4&48.6&51.8&53.4\\
\hline
Ours&\textbf{89.2}&\textbf{86.2}&\textbf{84.7}&\textbf{89.6}&\textbf{58.4}&\textbf{50.4}&\textbf{53.1}&\textbf{55.9}\\
\hline
\emph{IMP.} (\%)&1.48 &1.53 &3.17 & 1.01& 11.5& 3.57& 2.45& 4.12 \\
\hline
\end{tabular}}
\end{center}
%\vspace{-0.2cm}

\end{table*}

In this section, we apply our method to heterogeneous semi-supervised classification on two popular datasets: ACM and IMDB extracted by HAN~\cite{wang2019heterogeneous}. For a fair comparison, we follow the same experimental settings.

% \subsection{Comparison with Attention-based Method}
% \label{sec:hete_att}
\textbf{Meta-path based baseline:} Heterogeneous graph attention network (HAN) introduces hierarchical attention, including node-level and semantic-level attention. It can learn the importance between a node and its meta-path-based neighbours and the importance of different meta-paths for heterogeneous graph representation learning. 
%Like hop-aware attention methods in homogeneous graphs, the semantic-level attentions in HAN simultaneously scale the noise and information at the semantic level, which is not preferred. \\

\textbf{Ladder Aggregation at Semantic Level:} Since distinct meta-paths have different contributions to the target node representation, we propose to use \modelname~for semantic-level aggregation. In particular, with the prior knowledge of the ordinal importance of meta-paths (e.g., MDM is more important than MAM for movie type), we can allocate dimensions for them accordingly. Therefore, the hop-dimension relation function is used here for semantics-dimension relations, wherein more/less relevant semantic embeddings get higher/lower dimensions. Again, the only hyper-parameter in our method is the compression ratio $d$.
%we argue that assigning attention scores on different semantic-level embeddings can not disentangle messages to propagate them ineffectively. 

\textbf{Compare \modelname~with Attention-based Aggregation:} We use \emph{eight} kinds of meta-path combinations and compare with GAT\footnote{With GAT, we treat the heterogeneous meta-paths as homogeneous edges by summing up the adjacency matrices of different attributes.} and HAN. Experimental results on Top-1 accuracy are shown %on both the ACM dataset and IMDB dataset 
in Table~\ref{tab:hete_att}. From the results we can observe: (i) distinguishing heterogeneous paths is essential as the performance of GAT is always the worst; (ii) by allocating distinct dimensions for different semantic embeddings, the proposed \modelname~produces better results than state-of-the-art solution\footnote{More detailed analyses are shown in the appendix.} (i.e., HAN).

%\vspace{-0.2cm}
\subsection{Ablation Study}
\label{sec:homo_ab}
% \vspace{-0.1cm}
To analyze the impact of the hop-dimension function in Eq.~(\ref{eq:dim1}), we conduct experiments by varying two dominant hyper-parameters: the furthest hop $K$, the dimension compression rate $d$ and the aggregation methods among hops. We present the results on Citeseer.
%Although we design a NAS to explore for prior about these hyper-parameters, the ablation study in this section will help understand the impact of each hyper-parameter. Due to the space limitation, we only present the results on Citeseer.

\begin{table}[h]
\scriptsize
\begin{center}
\caption{Comparison of different compression rate $d$ under different furthest hop $K$ of the proposed \modelname.}
\label{tab:semi_de}
% \small
% \resizebox{\textwidth}{5mm}
{
\begin{tabular}{ l| c|c |c |c |c |c|c|c}
\hline
$K$ &$2$ &$3$&$4$ & $5$&$6$&$7$&$8$&$9$\\

\hline
$d$=$2$& 60.7&67.6&64.8&63.0&61.2&58.9&55.8&50.2\\
$d$=$0.5$&65.5&71.5&72.3&72.8&73.0&73.1&73.2&73.7\\
$d$=$0.25$& 68.8&72.9&73.3&73.9&73.6&73.5&73.4&73.0\\
$d$=$0.125$& \textbf{71.0}&\textbf{73.5}&\textbf{74.1}&74.0&74.2&\textbf{74.3}&\textbf{74.0}&72.8 \\
$d$=$0.0625$& 69.3&73.1&73.6&\textbf{74.7}&	\textbf{74.3}&\textbf{74.3}&73.8&\textbf{74.2}\\
$d$=$0.03125$ & 67.2&71.4&72.7&73.0&73.2&73.8&73.5&73.3\\
\hline

\end{tabular}}
\end{center}
% \vspace{-0.2cm}
\end{table}

\textbf{Impact of the largest hop $K$ and compression rate $d$.}

In Table~\ref{tab:semi_de}, the compression rate $d$ varies from $2$ to $0.03125$ ($1/32$) as comparison.
We can observe that
(i) by increasing the furthest hop $K$ with fixed $d$, the performance will increase to saturation when $d<1$.
When increasing $K$, more information can be aggregated from neighbors.
\modelname's ability to suppress noise by dimension compression facilitates the performance to saturation.
% The performance increase is because more information from neighbors is aggregated when increasing $K$.
% Moreover, \modelname~ can learn to suppress noise by dimension compression.
(ii) by decreasing the decay rate on fixed $K$, the performance first increases and then drops under most situations. The reason for increasing is that the decreased compression rate will map the distant nodes to a lower and suitable dimension space, suppressing the interference of noise in distant nodes. However, there is an upper bound for these improvements given $K$. When $d<0.0625$, the reduced dimension is too low to preserve the overall structural information, leading to worse performance in most cases. 
(iii) the effective rate $d$ is mainly on \{0.125, 0.0625\}, which can achieve better results for most $K$. If $K=5$ and $d=0.0625$, we obtain the best accuracy of 74.7\%. (iv) note that there are significant improvements with dimension compression comparing to dimension increase ($d=2$), which validates the effectiveness of the basic principle of dimension compression.
% \vspace{0.2cm}

\textbf{Impact of the aggregation operators.}

For aggregation method among hops, most existing solutions (e.g., GCN, SGC and GAT) mix the information from multiple hops, while Ladder-GNN disentangles and concatenates the information from multiple hops. 
To further demonstrate the effectiveness of the proposed method, we conduct an experiment to substitute concatenation with addition in Ladder-GNN. To accommodate the dimensional differences between features from different hops, we use zero paddings to fill those vacant positions before addition. As can be seen from Table~\ref{tab:agg}, concatenating features show consistently better results.

\begin{table}[h]
\scriptsize
% \vspace{-0.5cm}
\begin{center}
\caption{Comparison of different aggregation operators of the proposed \modelname.}
\label{tab:agg}
% \small
{
\begin{tabular}{ l| c|c |c}
\hline
Methods	&Citeseer&	Cora&	Pubmed\\
\hline
Concatenation&	\textbf{74.70$\pm$0.34}&\textbf{83.34$\pm$0.38}&\textbf{80.08$\pm$0.45}\\
Addition&73.15$\pm$0.40&80.06$\pm$0.51&	76.20$\pm$0.46\\
\hline
\end{tabular}}
\end{center}
% \vspace{-0.4cm}
\end{table}

%\vspace{-0.3cm}
\section{Conclusion}
% \vspace{-0.1cm}
In this work, we propose a novel, simple, yet effective ladder-style GNN aggregation scheme, namely \textit{Ladder-GNN}.
%, to tackle a long-standing challenge in GNN representation learning: how to improve the performance of nodes with low homophily by effectively extracting and integrating useful information embedded in long-distance nodes without sacrificing the performance of nodes with high homophily.   
%
We take a communication perspective for GNN representation learning, which motivates us to separate messages from different hops and assign different dimensions for them before concatenating them to obtain the node representation.
%
%The resulted representation facilitates extracting discriminative features effectively compared to exiting solutions. 
%
Experimental results on various semi-supervised node classification tasks show that the proposed solution can achieve state-of-the-art performance on most datasets. 
Particularly, at nodes with low homophily, the proposed solution outperforms existing aggregation schemes by a large margin.
This characteristic makes our solution a promising tool to enhance the generalization capability of existing GNNs, particularly in handling nodes with low homophily, which is a pain point of most existing GNNs.

\bibliography{arxiv}
\bibliographystyle{ieee_fullname}

\newpage
\appendix
\onecolumn

\section{Appendices}

In the appendix, we first detail our experimental settings, including datasets, our preprocessing procedure and other settings in Appendix~\ref{sec:ho_data}. 
%In Appendix~\ref{sec:app3}, we show additional ablation study. 
Then, we apply the proposed \modelname~on heterogeneous graph representation learning problem in Appendix~\ref{sec:app4}. Next, we discuss the limitations of our work in Appendix~\ref{sec:app5}. Finally, we present the broader impact of this work in Appendix~\ref{sec:app6}.
\appendix

\section{Experimental Settings}
Our experiments are conducted on a machine with an NVIDIA Tesla V100-SXM2-32GB GPU card. Most experiments take up a few hundred MBytes memory.

For the NAS experiment, the controller is a one-layer LSTM with $100$ hidden units. We train it with Adam optimizer. The learning rate is $3.5\times10^{-4}$, mini-batch is $128$, and the weights are randomly initialized, which is the same as GraphNAS~\cite{ijcai20gao} for a fair comparison. We train each sampled model $M$ with $150$ epochs, and the controller update step is $50$. For the conditionally progressive search algorithm, we filter hop-dimension combinations with $R_D(M)$ set as $0.8$, $0.7$, $0.78$ for \emph{Cora, Citeseer, Pubmed} datasets, respectively. Each model costs about $25$ seconds to get its reward $R_D(M)$ and hence it takes about $15$ days to get $50,000$ models. 

Considering computational efficiency, NAS may not be practical for graphs with a large amount of nodes and edges~\cite{hu2020open}. Therefore, except for the results shown in Sec.~\ref{sec:homo_nas}, all the other experiments are conducted with the proposed approximate hop-dimension relation function in Sec.~\ref{sec:3.4}. In these experiments, our training settings are the same as SGC~\cite{wu2019simplifying} with a fixed random seed. The learning rate is $0.2$ with Adam optimizer and the total number of epochs is $200$. All the hyper-parameters are listed in the scripts of our \emph{Code}.

\section{Homogeneous Data Description}
\label{sec:ho_data}
% \subsection{Data Description}
We use six datasets on semi-supervised node classification: Cora, Citeseer, Pubmed, OGB-Arxiv, and OGB-Products. The specifics of these datasets are listed in Table~\ref{tab:homo_data}.

\begin{table}[ht]
\begin{center}
\scriptsize
\caption{The statistics of the homogeneous datasets.}
\label{tab:homo_data}
{
\begin{tabular}{ l| c |c |c|c|c}
\hline
Dataset& Node& Edge&Feature Size&Train/Valid/Test&Class\\
\hline
Cora&2708&5429&1433&\multirow{3}{*}{20 per class/500/1000}&7\\
Citeseer&3327& 4732 &3703 &&6\\
Pubmed&19717&44338&500&&3\\
\hline
Flickr&7575 &239738 &12047 &180/360/540 &9\\
\hline
OGB-Arxiv&169343&1166243&128&54\%/18\%/28\%&40\\
OGB-Products&2449029&61859140&100&8\%/2\%/90\%&47\\

\hline
\end{tabular}}
\end{center}
% \vspace{-0.5cm}
\end{table}

\textit{Cora, Citeseer, Pubmed}\footnote{\url{https://github.com/kimiyoung/planetoid}} ~\cite{yang2016revisiting} are citation network datasets, where the nodes are papers and the edges are citation links. The node attributes are the bag-of-words features of papers, and nodes are divided into different areas. We follow the data preprocessing by \textit{SGC}\footnote{\url{https://github.com/Tiiiger/SGC}}.\\

% \textit{Flicker}\footnote{\url{https://github.com/mengzaiqiao/CAN}}~\cite{meng2019co} is a social network, where the nodes and the edges represent users and their relationships, respectively. The nodes are categorized into nine classes according to the interest groups of users. We follow the data preprocessing by \textit{AM-GCN}\footnote{\url{https://github.com/zhumeiqiBUPT/AM-GCN}}.\\

\textit{OGB-Arxiv}\footnote{\url{https://ogb.stanford.edu/docs/nodeprop}}~\cite{hu2020open} is a citation network between all Computer Science (CS) arXiv papers indexed by MAG~\cite{wang2020microsoft}, where nodes are arxiv papers and edges are citation links. The node feature comes from a 128-dimensional feature vector generated by averaging the embeddings of words in its title and abstract. We follow the data preprocessing by \textit{AGD}\footnote{\url{https://github.com/skepsun/adaptive$\_$graph$\_$diffusion$\_$networks$\_$with$\_$hop-wise$\_$attention}}.\\

\textit{OGB-Products}\footnote{\url{http://manikvarma.org/downloads/XC/XMLRepository.html}}~\cite{hu2020open} is an Amazon product co-purchasing network. Nodes represent products sold on Amazon, and edges between two products indicate that the products are purchased together. We follow previous works~\cite{chiang2019cluster,hu2020open} to process node features and target categories. Specifically, node features are obtained by extracting bag-of-words features from the product descriptions followed by a Principal Component Analysis, and its dimension is reduced to 100. We follow the data preprocessing by \textit{SAGN}\footnote{\url{https://github.com/skepsun/SAGN$\_$with$\_$SLE}}.

\section{Heterogeneous Data Description}
\label{sec:app4}

The specifics of the corresponding datasets are listed in Table~\ref{tab:hete_data}. 

% \subsection{Heterogeneous Semi-supervised Node Classification}
% \label{sec:hete}

\begin{table*}[htbp]
\begin{center}
\scriptsize
\caption{The statistics of the heterogeneous datasets}
\label{tab:hete_data}
%\resizebox{\textwidth}{5mm}
{
\begin{tabular}{ l| c |c |c |c|c|c|c |c|c|c}
\hline
Dataset& Relations (A-B)&Node A&Node B&Edge (A-B)&Feature&Training&Validation&Test&Class&Meta Path\\
\hline
\multirow{3}{*}{ACM}&Paper-Author&\multirow{3}{*}{3025}&5835&16153&\multirow{3}{*}{1830}&\multirow{3}{*}{600}&\multirow{3}{*}{300}&\multirow{3}{*}{2125}&\multirow{3}{*}{3}&PAP\\
&Paper-Subject&&56&1106893&&&&&&PSP\\
&Paper-Term&&3025&4576809&&&&&&PTP\\
\hline
\multirow{3}{*}{IMDB}&Movie-Actor&\multirow{3}{*}{4780}&5841&51395 &\multirow{3}{*}{1232} &\multirow{3}{*}{300}&\multirow{3}{*}{300}&\multirow{3}{*}{2687} &\multirow{3}{*}{3}&MAM\\
&Movie-Director&&2269&12899&&&&&&MDM\\
&Movie-Year&&4780&409316&&&&&&MYM\\

\hline
\end{tabular}}
\end{center}
% \vspace{-0.5cm}
\end{table*}

\textit{ACM}\footnote{\url{http://dl.acm.org/}}~\cite{wang2019heterogeneous} is sourced from the ACM database, where the nodes are papers (published in KDD, SIGMOD, SIGCOMM, MobiCOMM, and VLDB). The heterogeneous graph consists of three kinds of meta-path sets \{PAP, PSP, PTP\}. Each node represents the bag-of-words features of keywords, and these nodes are to be categorized as three classes (Database, Wireless Communication, Data Mining). We follow the data preprocessing by \textit{HAN}\footnote{\url{https://github.com/Jhy1993/HAN/}}.

\textit{IMDB}\footnote{\url{https://www.imdb.com/}}~\cite{wang2019heterogeneous} is a subset of the online IMDB database extracted by HAN~\cite{wang2019heterogeneous}, where the nodes are movies classified into three types (Action, Comedy, Drama). The heterogeneous graph consists of three kinds of meta-path sets \{MAM, MDM, MYM\}. The features of movies correspond to the elements of plots. We follow the data preprocessing by \textit{HAN}\footnote{\url{https://github.com/Jhy1993/HAN/}}.

\section{Results of High-order Aggregation}
\label{sec:high}
We take the combinations of two aggregation framework among nodes with our proposed~\modelname~as examples.

\begin{figure*}[h!]	
	\subfigure[Cora Dataset] %第一张子图
    	{
    		\begin{minipage}{7.1cm}
    			\centering          %子图居中
    			\includegraphics[scale=0.35]{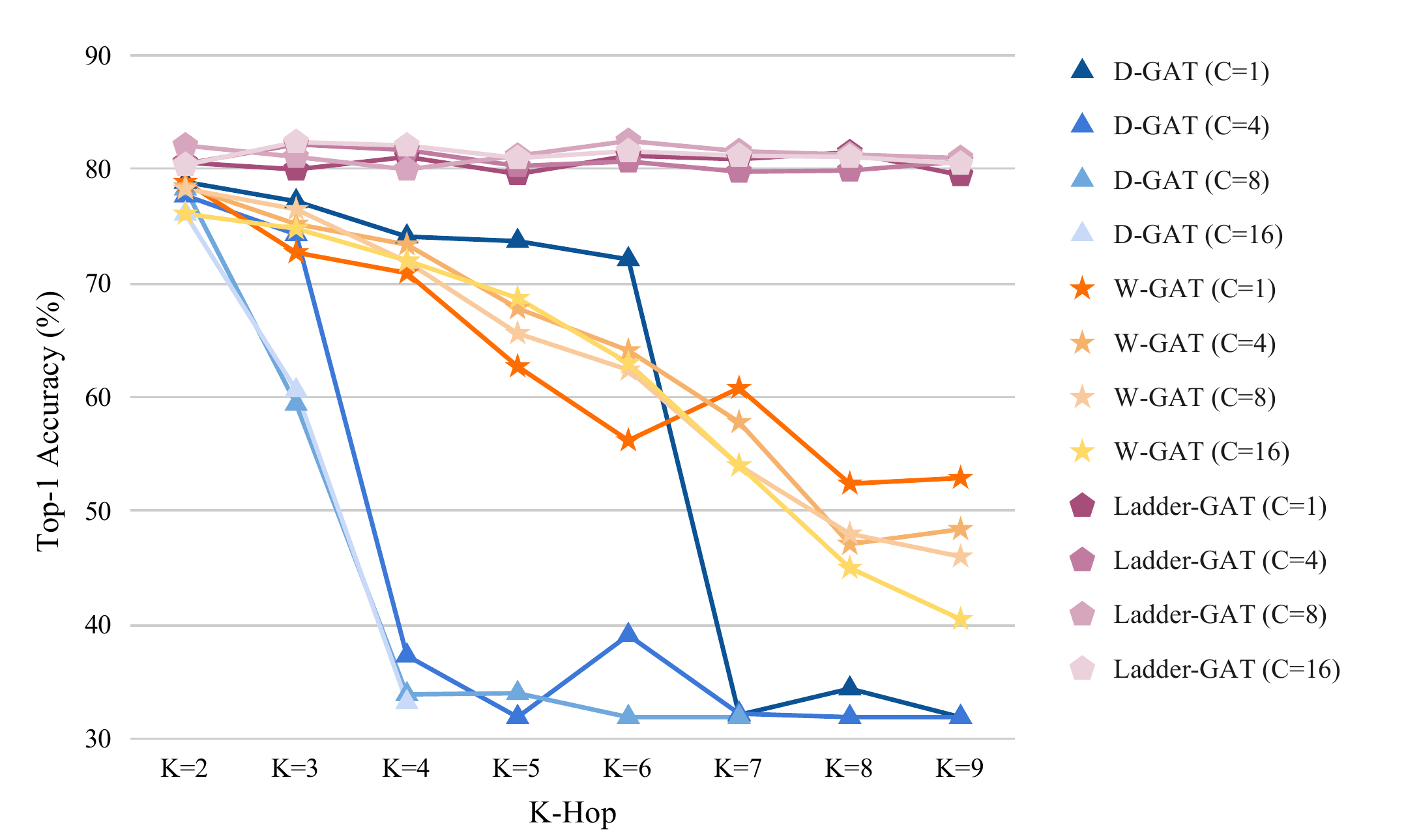}   
    		\end{minipage}
    	}
	\subfigure[Citeseer Dataset] %第二张子图
    	{
    		\begin{minipage}{4.4cm}
    			\centering      %子图居中
    			\includegraphics[scale=0.35]{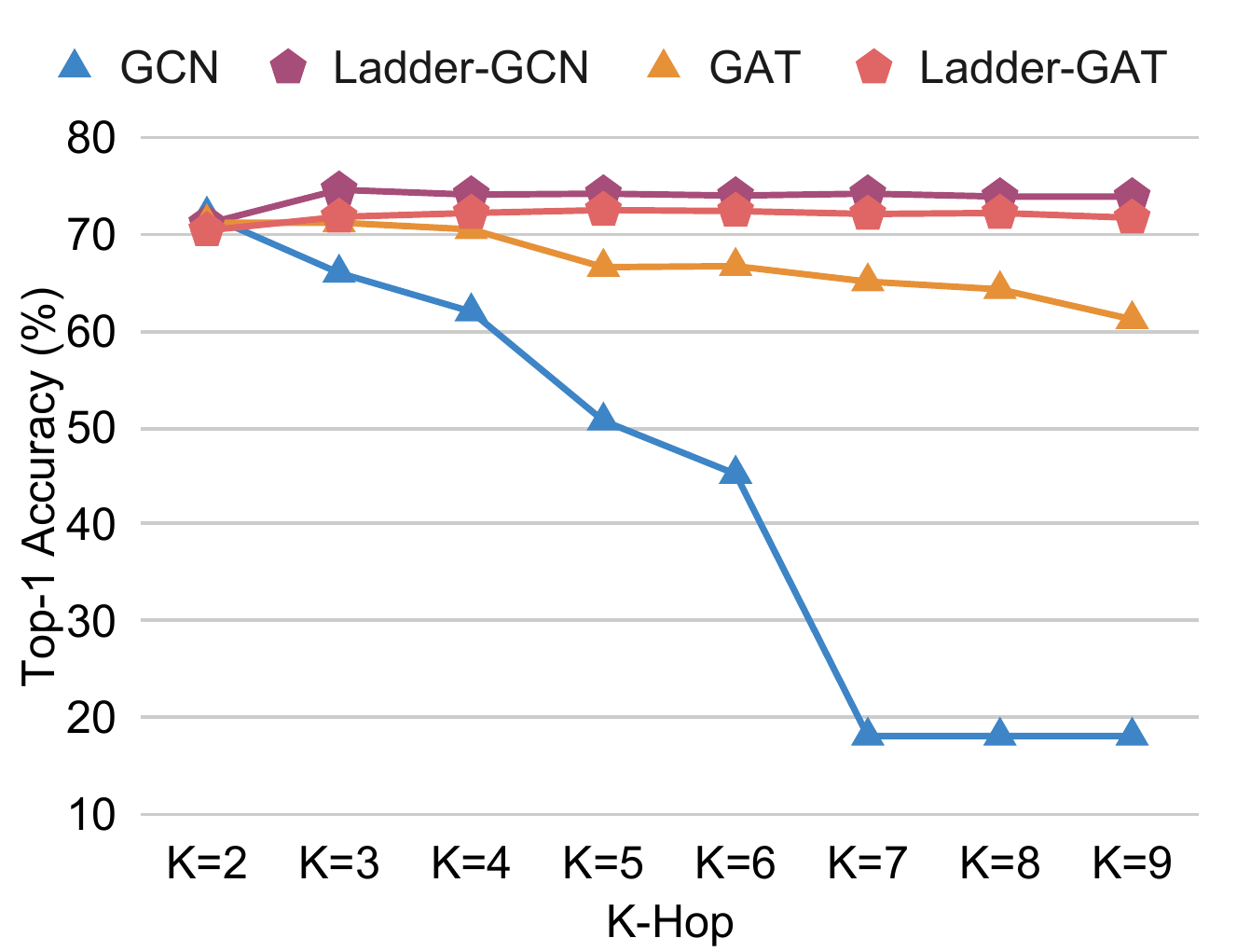}   
    		\end{minipage}
    	}
	\subfigure[Pubmed Dataset] %第二张子图
    	{
    		\begin{minipage}{4cm}
    			\centering      %子图居中
    			\includegraphics[scale=0.35]{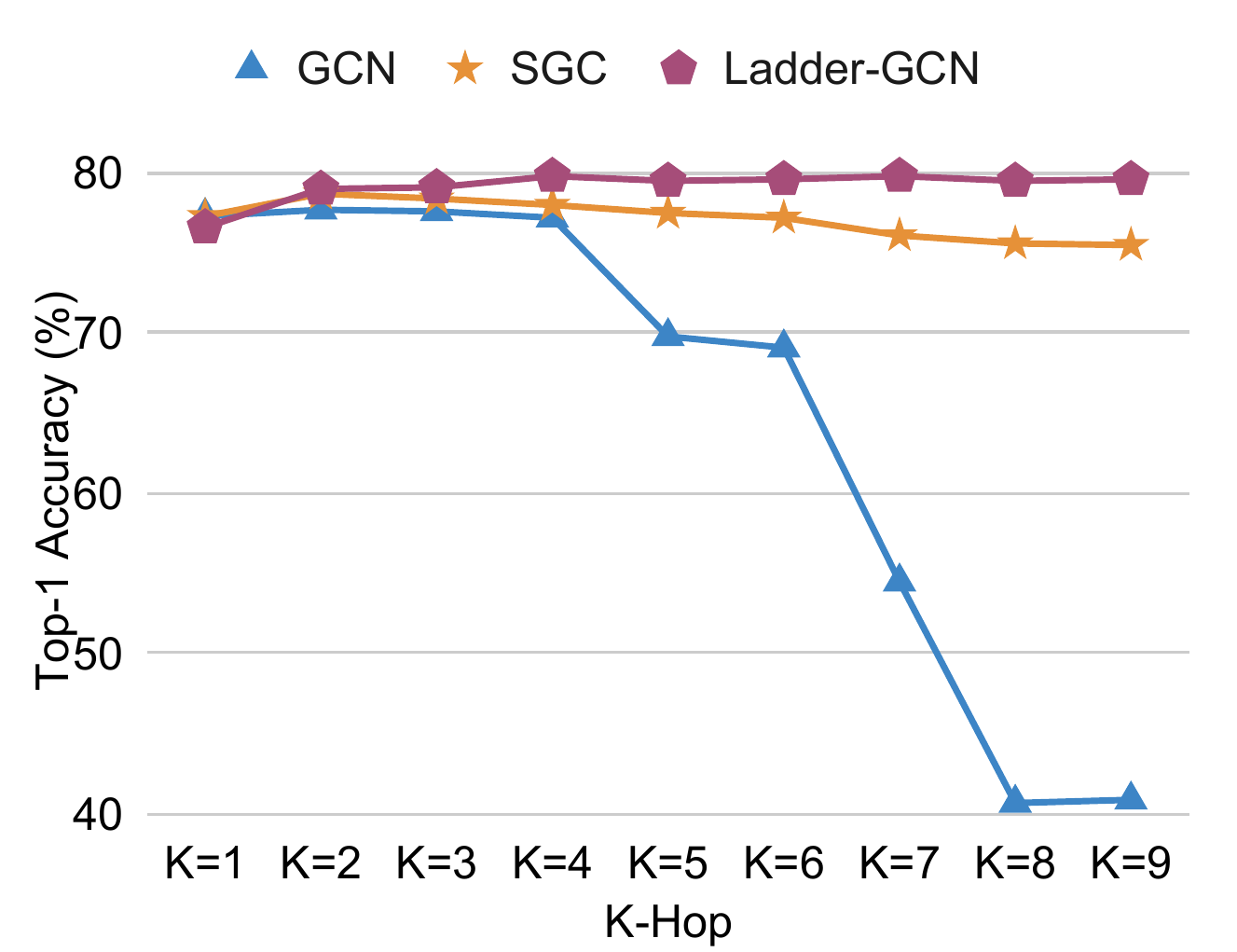}   
    		\end{minipage}
    	}
	\vspace{-0.2cm}
	\caption{Accuracy comparison of different hop-level aggregation methods on different datasets.} %  %大图名称
	\label{fig:cora_vis}  %图片引用标记
\end{figure*}
%GAT~\cite{velivckovic2017graph} tends to distinguish the relationships among direct nodes, showing better performance than GCN in many complicated graphs. 
%\textbf{Result of high-order aggregation of GAT.} 
For GAT, we explore whether GAT itself can aggregate higher-order neighbors effectively. Hence, we use eight multi-heads and four kinds of channel sizes \{1, 4, 8, 16\} for each head in their self-attention scheme. To aggregate $K$ hop neighbors, we set two kinds of baselines with a \emph{deeper} structure (stacking $K$ layers), named as \emph{D-GAT} in {\color[RGB]{56,108,176}{blue}} lines and a \emph{wider} (one layer computes and aggregates multiple-order neighbors) network \emph{W-GAT} in {\color[RGB]{217,95,2}{orange}} lines. In Figure~\ref{fig:cora_vis}(a), we demonstrate the accuracy of \emph{D-GAT} and \emph{W-GAT} with ours in {\color[RGB]{84,39,136}{purple}} line. As we can observe, the \emph{D-GAT} will suffer from \emph{over-smoothing} problems suddenly ($K$=$4$), especially for the larger channel size. Moreover, the performance of \emph{W-GAT} will degrade gradually, due to the \emph{over-fitting} problems with more parameters. Thus, both of the two aggregations drop their performance as hops increase. On the contrary, the proposed \emph{Ladder-GAT} is robust to the increasing of hops since the proposed~\modelname~can relieve the above problems when aggregating high-order neighbors.

%to boost the performance and robustness when aggregating high-order neighbors

%\textbf{Results of high-order aggregation of GCN.}
%Through the introduction of the Sec.~\ref{sec:comb}, we further combine ladder aggregation into GCN as Ladder-GCN. To explore their effectiveness, 
In Figure~\ref{fig:cora_vis}(b), we compare the original GCN and GAT methods with the proposed \emph{Ladder-GCN} and \emph{Ladder-GAT} on Citeseer dataset. We can obtain similar observations. As the high-order neighbors are aggregated, GCN and GAT encounter performance degradation, while our framework can boost the performance and relieve the potential over-smoothing issue. 

In Figure~\ref{fig:cora_vis}(c), we compare GCN, SGC with \emph{Ladder-GCN}, where the low-order $L$ hops are aggregated without dimension reduction and compress the dimensions of hop $k$ (L $<$ $k$ $\leq$ K). The {\color[RGB]{84,39,136}{purple}} line is set as an interesting setting: only compress the last $K_{th}$ hop to a lower dimension $32$ and $L$=$K$-$1$. Under the same horizontal coordinate $K$ hops, \emph{Ladder-GCN} achieves consistent improvement compared with both GCN and SGC (in {\color[RGB]{217,95,2}{orange}} line). The reasons behind this result is that the parameters updated in SGC are affected by (i) the decrescent information-to-noise ratios within distant hops
and (ii) over-squashing phenomena~\cite{alon2020bottleneck} -- information from the exponentially-growing receptive field is compressed by fixed-length node vectors, and causing it is difficult to make $k_{th}$ hop neighbors play a role. This observation suggests that compressing the dimension on the last hop can mitigate the \emph{over-squashing problem} in SGC, which consistently improves the performance of high-order aggregation.

\section{Analyses the Differences between \modelname~ and Attention-based Aggregation} 
\begin{figure*}[htbp]	
% 	\subfigure[] %第一张子图
	{
		\begin{minipage}{5.2cm}
			\centering          %子图居中
			\includegraphics[scale=0.37]{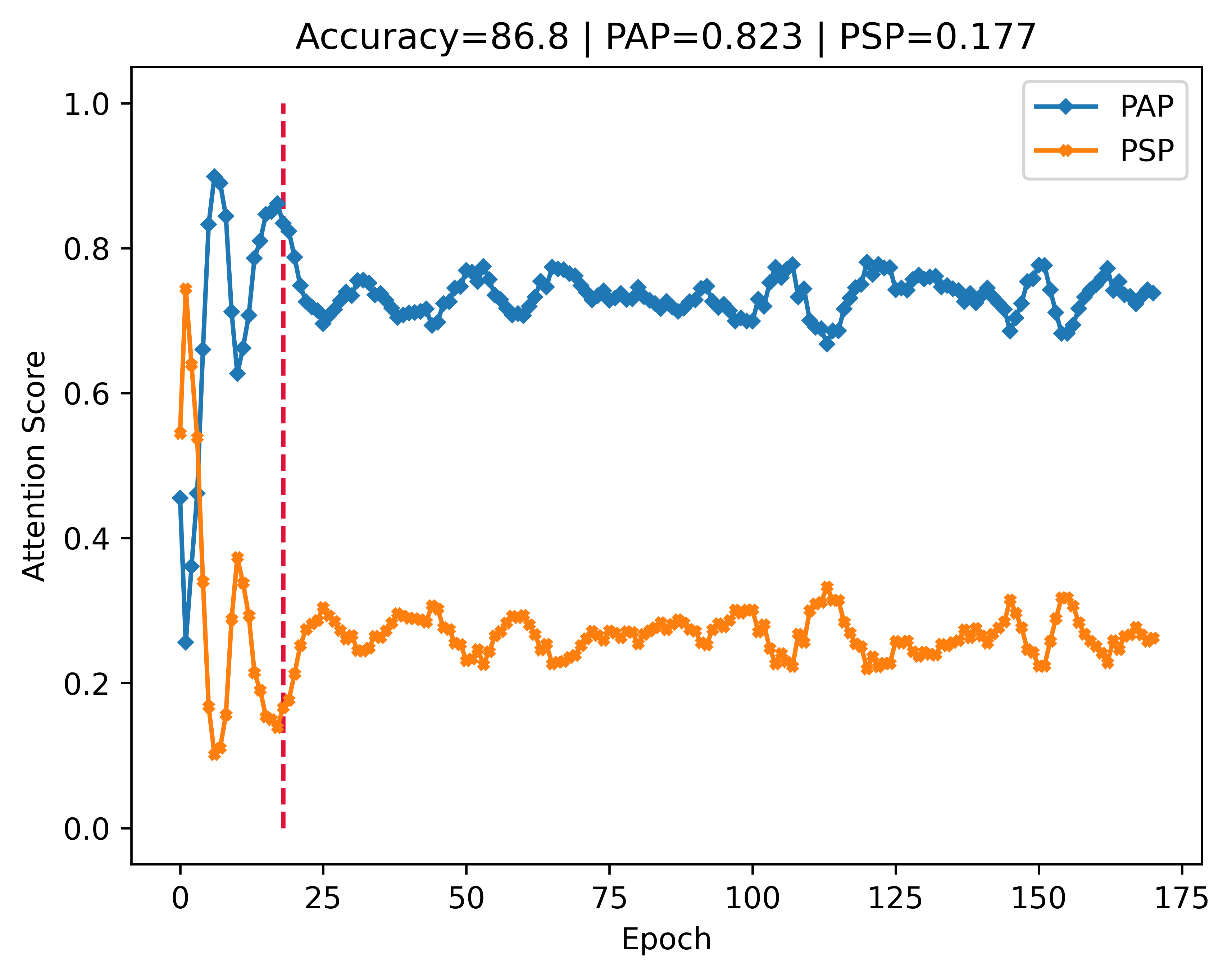}   %以pic.jpg的0.4倍大小输出
		\end{minipage}
	}
	%\hspace{0.2cm}
% 	\subfigure[] %第二张子图
	{
		\begin{minipage}{5.2cm}
			\centering      %子图居中
			\includegraphics[scale=0.37]{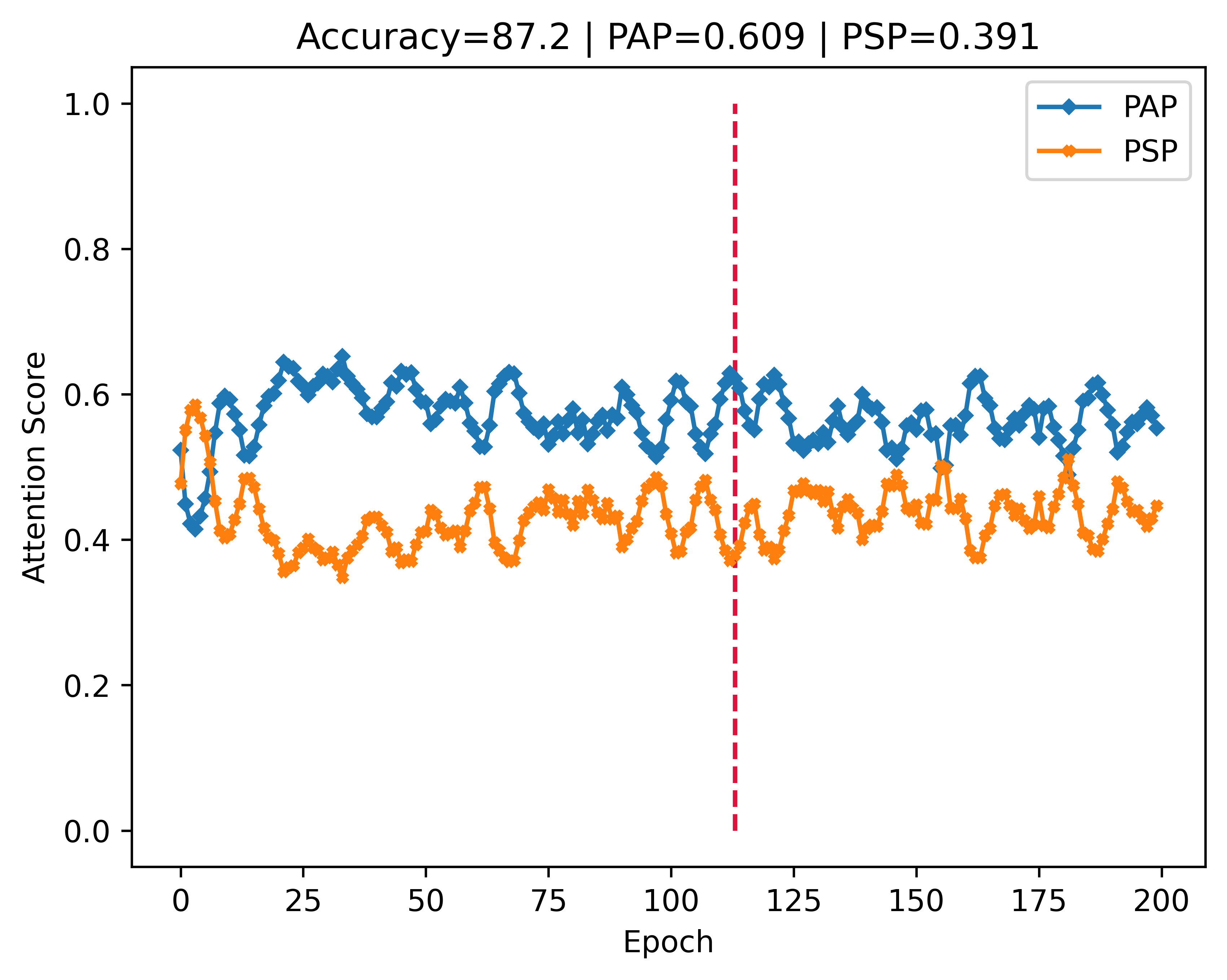}   %以pic.jpg的0.4倍大小输出
		\end{minipage}
	}
	%	\hspace{0.2cm}
% 	\subfigure[] %第二张子图
	{
		\begin{minipage}{5.2cm}
			\centering      %子图居中
			\includegraphics[scale=0.37]{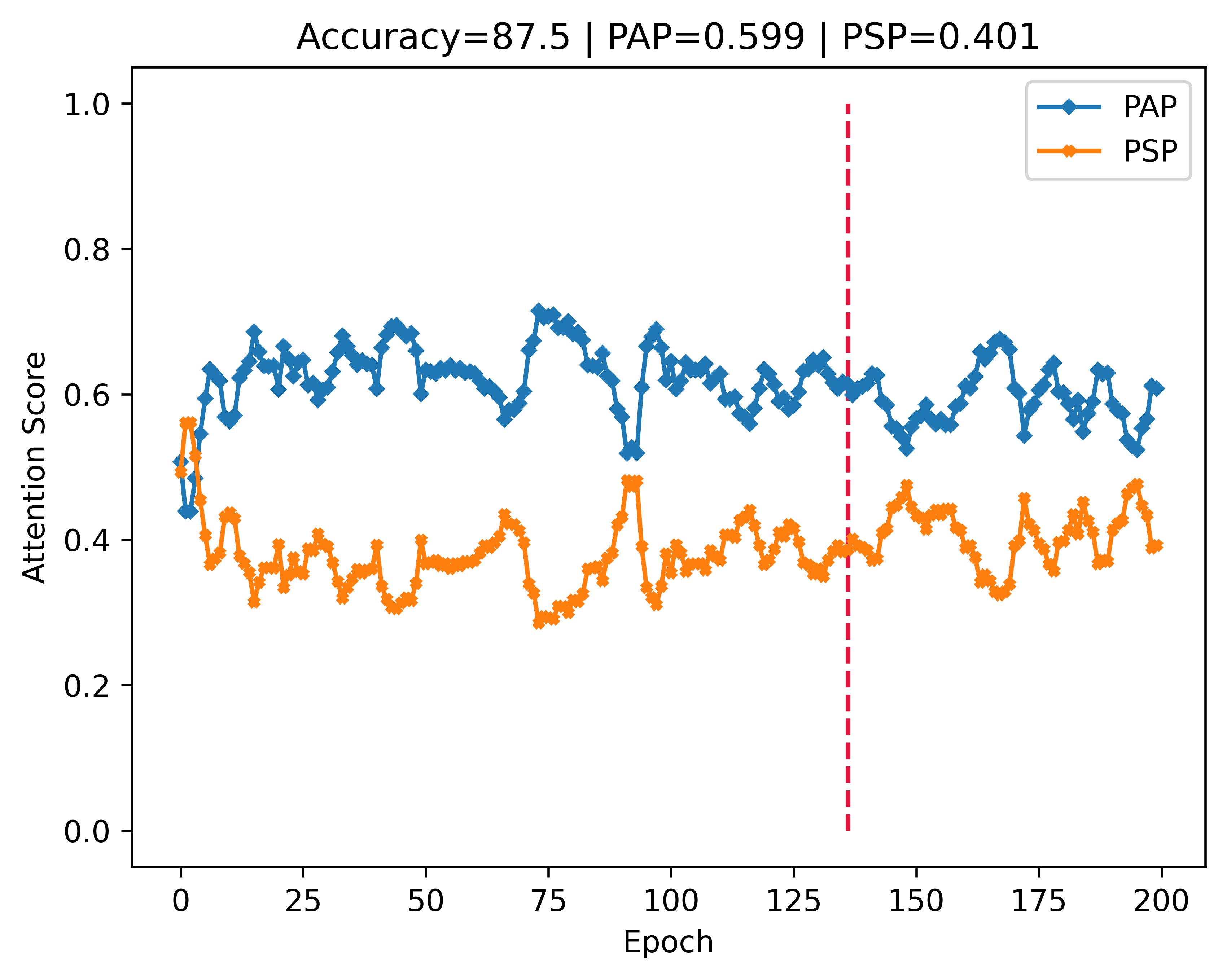}   %以pic.jpg的0.4倍大小输出
		\end{minipage}
	}
	\vspace{-0.4cm}
	\caption{Comparison of three kinds of \emph{semantic attention scores} between \{PAP, PSP\} meta-path from HAN models with similar accuracy. The X-axis illustrates training epochs, while Y-axis shows semantic attention scores. The red dotted line in each subgraph indicates the epoch obtaining the best validation set accuracy.} %  %大图名称
	\label{fig:han_att}
	\vspace{-0.2cm}
	 %图片引用标记
\end{figure*}

Based on the comparisons in Section~\ref{sec:hete_exp} of heterogeneous graph representation learning, 
we further study the semantic attention scores learned in HAN. In Figure~\ref{fig:han_att}, we demonstrate three kinds of semantic attention scores between two meta-paths in the ACM dataset from different self-attention channels from node level, e.g., \{4, 8, 16\} of each head and the head is $4$ here. We find that (i) although these three models obtain similar accuracy, the patterns of their attention scores are quite different, and the scores of the best accuracy are also different, for instance, ~\{PAP=0.823, PSP=0.177\} for the first model while \{PAP=0.609, PSP=0.391\} for the second model; (ii) the learnable attention scores are changed by epochs, which can be vulnerable with training strategies; (iii) if only a score is multiplied over the whole semantic features, then both useful and useless information of the features will be scaled simultaneously. Therefore, the information-to-noise ratio will not be changed. Thus, the attention-based methods tend to obtain worse performance. Compared with these methods, our method shows consistent improvements and proves its effectiveness.

\section{Comparison with Existing Work}
\label{sec:homo_results}

Table~\ref{tab:comp} shows results on \modelname~with other two groups of existing methods, such as general GNNs and GNNs with modified graph structures, in terms of Top-1 accuracy (\%) on the most-used datasets. Meanwhile, Table~\ref{tab:node} compares \modelname~with existing methods on larger graphs. As a new hop-aware aggregation, our method can achieve the state-of-the-art performance in most cases by improving the hop-aware aggregation method of GCN and GAT, and the proposed method can improve GCN and GAT consistently. Specifically, Ladder-GCN surpasses 1.5\% with \emph{GXN} on Citeseer, 0.5\% with \emph{DisenGCN} on Pubmed, respectively. Our method can still show superiority on Citeseer, showing its effectiveness to handle those graphs with lower information-to-noise ratio. Although there are kinds of variants on GNNs, we verify the potential of a better hop-aware aggregation method. Hence, applying the proposed method to other structure modification and aggregation within one hop will be a future direction.

\begin{table}[htbp]
\scriptsize
\caption{The comparison of General GNNs.}
\label{tab:comp}
\begin{center}
{
\begin{tabular}{ l |l|c |c|c}
\hline
\multicolumn{2}{c|}{Method}&Cora&Citeseer&Pubmed\\
\hline
\multirow{13}{*}{\rotatebox[origin=c]{90}{General GNNs-based methods}}
&ChebNet~\cite{defferrard2016convolutional}&81.2&69.8&74.4\\
&GCN~\cite{kipf2016semi}&81.5&70.3&79.0\\
&GraphSage~\cite{hamilton2017inductive}&81.3&70.6&75.2\\
&GAT*~\cite{velivckovic2017graph}&78.9&71.2&79.0\\
&GIN~\cite{xu2018powerful}&77.6&66.1&77.0\\
&JKNet~\cite{xu2018representation}&80.2&67.6&78.1\\
&SGC~\cite{wu2019simplifying}&81.0&71.9&78.9\\
&APPNP~\cite{klicpera2018predict}&81.8&72.6&79.8\\
&ALaGCN~\cite{xie2020gnns}&82.9&70.9&79.6\\
&GraphHeat~\cite{xu2020graphheat} &\underline{83.7}&72.5&\underline{80.5}\\
&MCN~\cite{lee2018higher}&83.5&73.3&79.3\\
&DisenGCN~\cite{ma2019disentangled}&\underline{83.7}& 73.4& \underline{80.5}\\
&SEGNN~\cite{dai2021towards}&80.4&73.8&80.0\\
&FAGCN~\cite{bo2021beyond}&\textbf{84.1}&72.7&79.4\\
&S$^{2}$GC~\cite{zhu2021simple}&83.5&73.6&80.2\\
\hline
\multirow{6}{*}{\rotatebox[origin=c]{90}{Structure}}
&DropEdge-GCN~\cite{rong2019dropedge}&82.0&71.8&79.6\\
&DropEdge-IncepGCN~\cite{rong2019dropedge}&82.9&72.7&79.5\\
&AdaEdge-GCN~\cite{chen2020measuring}&82.3&69.7&77.4\\
&PDTNet-GCN~\cite{luo2021learning}&82.8&72.7&79.8\\
&GXN~\cite{zhao2020data} &83.2&73.7&79.6\\
&SPAGAN~\cite{yang2021spagan}&83.6&73.0&79.6\\
\hline
&Ladder-GAT&82.6&\underline{73.8}&\textbf{80.6} \\
&Ladder-GCN&83.3&\textbf{74.7}&80.0 \\
\hline
\end{tabular}}
\end{center}

\end{table}

\begin{table}[htbp]
% \small
\scriptsize
\caption{The comparison with existing methods on Flicker, OGB-Arxiv and OGB-Products datasets.}
% \vspace{-5mm}
\begin{center}
{
\begin{tabular}{l|c|c|c}
\hline
\multicolumn{1}{c|}{Methods}  & Flicker & OGB-Arxiv & OGB-Products \\ \hline
% ChebNet~\cite{defferrard2016convolutional}      & 23.3    & -     & -        \\ 
GCN~\cite{kipf2016semi}          & 41.1    & 71.7  & 75.6     \\ 
GraphSage~\cite{hamilton2017inductive}    & 57.4    & 71.5  & 78.3     \\ 
GAT*~\cite{velivckovic2017graph}          & 46.9    & \underline{73.6}  & \underline{79.5}     \\ 
JKNet~\cite{xu2018representation}        & 56.7    & 72.2  & -        \\ 
SGC~\cite{wu2019simplifying}          & 67.3    & 68.9  & 68.9     \\ 
APPNP~\cite{klicpera2018predict}        & -       & 71.4  & -        \\ 
DropEdge-GCN~\cite{rong2019dropedge} & 61.4    & -     & -        \\ 
MixHop~\cite{abu2019mixhop}       & 39.6    & -     & -        \\ 
S$^{2}$GC~\cite{zhu2021simple}         & -       & 72.0  & 76.8     \\ 
\hline 
Ladder-GCN   & \textbf{73.4}    & 72.1  & 78.7     \\ 
Ladder-GAT   & \underline{71.4}    & \textbf{73.9} & \textbf{80.8}     \\ 
\hline
\end{tabular}}
\end{center}
\label{tab:node}
\end{table}

\section{Limitations and Future Work}
\label{sec:app5}

The proposed \modelname~focuses on hop-level aggregation and the dimensionality of high-order neighbors is reduced with our hop-dim relation function. The obtain high-quality node representations, \modelname~relies on the quality of the given graphs for learning. That is, for node classification problems, the homogeneity ratio of low-order neighbors is higher than that of high-order neighbors for \modelname~to perform well. While this is true in most cases, it would be better to jointly optimize the structure of the graph and \modelname, especially for large graphs with a huge amount of nodes and complex relations. We plan to investigate this joint opimiztion problem in our future work.

\section{Broader impact}
\label{sec:app6}

Graph neural networks(GNNs) have been widely used in various real-world applications ranging from chemo- and bioinformatics to recommendation systems and social network analysis. Although our method can boost the development of the GNNs, we also notice that there should be a negative impact of this technique on protecting personal privacy and security. For instance, customers’ behaviour may be easily predicted based on their historical relevant hobbies, which provides a convenient way for unscrupulous business people to send spam information. Therefore, we call on researchers to cultivate a responsible AI-ready culture in technology development and avoid the abuse of technologies.

\end{document}